\documentclass[bachelor,english]{hgbthesis}

\usepackage[utf8]{inputenc}
\usepackage{amsfonts}
\usepackage{amsmath}
\usepackage{bm}
\usepackage{listings}
\usepackage[toc,page]{appendix}

\graphicspath{{images/}}    
\logofile{logo}				
\title{Using Deep Reinforcement Learning for the Continuous Control of Robotic Arms}
\author{Winfried Lötzsch}
\programname{B. Sc. Informatik}
\placeofstudy{TU Chemnitz}
\dateofsubmission{2017}{09}{18}	

\semester{Summer semester 2017} 	
\advisor{Dr. habil. Julien Vitay, Prof. Dr. Fred Hamker}

\newcommand\numberthis{\addtocounter{equation}{1}\tag{\theequation}}

\DeclareMathOperator*{\argmin}{arg\,min}
\DeclareMathOperator*{\argmax}{arg\,max}
\numberwithin{equation}{section}

\newcommand{\pluseq}{\mathrel{{+}{=}}}
\newcommand{\minuseq}{\mathrel{{-}{=}}}

\begin{document}

\frontmatter

\maketitle
\tableofcontents

\chapter{Abstract}
Deep reinforcement learning enables algorithms to learn complex behavior, deal with continuous action spaces and find good strategies in environments with high dimensional state spaces. With deep reinforcement learning being an active area of research and many concurrent inventions, we decided to focus on a relatively simple robotic task to evaluate a set of ideas that might help to solve recent reinforcement learning problems. We test a newly created combination of two commonly used reinforcement learning methods, whether it is able to learn more effectively than a baseline. We also compare different ideas to preprocess information before it is fed to the reinforcement learning algorithm. The goal of this strategy is to reduce training time and eventually help the algorithm to converge. The concluding evaluation proves the general applicability of the described concepts by testing them using a simulated environment. These concepts might be reused for future experiments.

\mainmatter

\chapter{Introduction}
Artificical intelligence is an interdisciplinary field, that receives input from many sources. A recent renaissance of neural networks in machine learning is mainly motivated by technical considerations. Especially the availability of large data sets and massive computational power enabled deep network structures to solve problems, that were considered infeasible before (\cite{silver2016mastering}). The term \textit{deep learning} generally refers to training any neural network structure with multiple computing layers. As more layers, that compute non-linear functions of previous layers, are added, the complexity of the network rises and solutions to more difficult tasks can be found. It has been shown, that even if using a shallow neural network model is possible, deeper architectures are often much more efficient, as they can be trained faster and consume less memory (\cite{goodfellow2016deep}).

An interesting implication of the success of deep neural networks is their application to reinforcement learning. In contrast to other machine learning techniques, reinforcement learning can be used to learn complex strategies or behavior. Applications of reinforcement learning range from optimization problems to robotics and complex control problems (\cite{abbeel2007application}). Deep reinforcement learning incorporates the insights of deep learning into reinforcement learning and thereby effectively overcomes many common limitations like being able to act, when the received data is multi-dimensional and complex.

This thesis shortly motivates the use of reinforcement learning in chapter \ref{ch:typesOfLearning} by comparing it very generally to other machine learning approaches. Chapter \ref{ch:variantsOfRL} states common limitations of reinforcement learning in practice and describes various ways to overcome these limitations. The main purpose of this chapter is to show, that deep learning is the answer to many open questions in the field. Several ways to benefit from deep learning under the given circumstances will be discussed and compared.

Finding solutions to complex problems with deep reinforcement learning can still be slow and there is sometimes no gurantee for a standard algorithm to succeed solving a previously unseen task, especially if there is no time to carefully choose the hyperparameters of the algorithm. Chapter \ref{ch:extensionsToRl} describes ideas to train models that preprocess the available information in advance and thus simplify the reinforcement learning task. We call this strategy pretraining.

One goal of this thesis is to investigate the concepts of chapter \ref{ch:extensionsToRl} and their practical applicability to algorithms of chapter \ref{ch:variantsOfRL}. We also combine DDPG (deep deterministic policy gradient; \cite{lillicrap2015continuous}) and asynchronous methods (\cite{mnih2016asynchronous}), which both are commonly used reinforcement learning methods, to form two new reinforcement learning algorithms, which will be investigated. We call these algorithms distributed and asynchronous DDPG. The intention of combining two inventions, that were recently succesful in the field, is to combine their respective benefits. Asynchronous methods enable fast training and generalize well to unseen situations. The DDPG algorithm is a good choice for learning a continuous action space. This means, that a discrete choice between several options is not satisfactory, but complex continuous actions need to be generated for a correct behavior. Chapter \ref{ch:methods} introduces these algorithms, but also describes a simulated robotic environment that was used for all evaluations and some interesting implementation details. We implemented the distributed and asynchronous DDPG algorithms and reused a DDPG implementation as baseline.

Chapter \ref{ch:experimentalResults} explains the experimental setups and states the obtained results. We evaluate all variants of DDPG and compare the respective scores, but also combine the DDPG baseline with differently pretrained models for comparing the quality of various pretraining techniques. Chapter \ref{ch:discussion} concludes with a discussion.

\chapter{Different Types of Learning}
\label{ch:typesOfLearning}

\section{Supervised Learning}
A large number of machine learning tasks like regression, classification or pattern recognition aim to approximate a function \(\bm{y}=f(\bm{x})\) given a set of training examples \((\bm{x}^{(i)},\bm{y}^{(i)})\). These tasks can be viewed as approximating the probability distribution of the given data \(p_{data}(\bm{y}|\bm{x})\). The formulation is taken from \textcite{goodfellow2016deep}. While other formulations are possible, estimating a probability distribution is very general and has the pleasing property that many learning algorithms can be derived easily, for example using maximum likelihood estimation to match the distribution of the model to the data distribution.

As an example, performing maximum likelihood estimation in a simple supervised learning model with only one output is equal to minimizing the mean squared error (MSE) between the outputs of the model and the targets \(y^{(i)}\), if the distribution of the model is assumed to be gaussian with fixed variance and mean \(\hat{y}\) specified by the model with learnable parameters \(\bm{\theta}\) (\cite{goodfellow2016deep}). Maximum likelihood estimation is performed by maximizing the conditional log-likelihood over the training examples \((\bm{x}^{(i)},y^{(i)})\) to turn the possibly large product of probabilities \(\prod_{i}p_{model}(y^{(i)}|\bm{x}^{(i)};\bm{\theta})\) into a sum and overcome issues like numerical underflow. The training examples are assumed to be i.i.d. to make this a valid conversion.

\begin{equation}
\hat{y}^{(i)}=f_{model}(\bm{x}^{(i)};\bm{\theta})
\end{equation}
\begin{equation}
p_{model}(y|\bm{x};\bm{\theta})=\mathcal{N}(y;\hat{y}, \sigma^2)
\end{equation}
\begin{align}
\bm{\theta}_{ML}=&\argmax_{\bm{\theta}}\sum_{i}\log p_{model}(y^{(i)}|\bm{x}^{(i)};\bm{\theta})\\
=&\argmax_{\bm{\theta}}\sum_{i}\log(\frac{1}{\sqrt{2\pi\sigma^2}}e^{-\frac{(y^{(i)}-\hat{y}^{(i)})^2}{2\sigma^2}})\\
=&\argmax_{\bm{\theta}}\sum_{i}-\log\sigma-\frac{1}{2}\log(2\pi)-\frac{(y^{(i)}-\hat{y}^{(i)})^2}{2\sigma^2}
\end{align}

Removing all terms that do not depend on the parameters \(\bm{\theta}\) yields a formula very similar to MSE. Usually it is not possible to directly estimate \(\bm{\theta}_{ML}\) due to computational limitations. Therefore, it is necessary to perform gradient descent to optimize the parameters step by step.
\begin{equation}
\bm{\theta}_{ML}=\argmin_{\bm{\theta}}\sum_{i}(y^{(i)}-\hat{y}^{(i)})^2
\end{equation}

\section{Unsupervised Learning}
When there is no output specified, machine learning models can still approximate a distribution  \(p_{data}(\bm{x})\) and thus learn to find structures inherent in the data. This can be useful for tasks like denoising, clustering or pretraining some parameters of a model for a supervised learning task. \textcite{vincent2008extracting} and \textcite{kingma2013auto} provide examples of successfully applied unsupervised learning.

\section{Reinforcement Learning}
\label{sec:rl}

Many real world applications like robotic tasks or playing video games require learning to perform a series of actions \(a_t\), for instance pressing buttons or moving a robotic arm to a target by controlling motors, while the state of the environment \(s_t\) can be observed between the actions often forming a trajectory over time: \(\tau=\{s_1,a_1,s_2,a_2...s_T\}\). Each complete trajectory starting from an initial state forms an \textit{episode}, which can be ended after a predefined number of time steps or after reaching a terminal state. Besides trajectory centric or episodic reinforcement learning, there are \textit{continuing} environments that do not use trajectories of fixed length, but visit every possible state infinitely often with a certain probability (\cite{sutton1998reinforcement}). A common property of reinforcement learning tasks is that there is no previously known solution to the problem but an implicit aim in form of a reward, which could be the score in a video game or a distance measure stating how well a robot performed the task of moving to a specific position. Learning from this kind of signal is biologically plausible, because some brain areas like the basal ganglias are supposed to follow the same strategy (\cite{doya2000complementary}).

Reinforcement learning enables intelligent algorithms to learn, when the objective is not to directly model a specific probability distribution like discussed above, but to maximize a scalar reward signal \(r_{t+1}=r(s_t,a_t)\) assigned to each pair of state and action over time. The reward at each time step depends on both the current state and the chosen action. At each time step the action \(a_t\) is sampled from a policy function \(p(a_t)=\pi_\theta(a_t|s_t)\) with learned parameters \(\theta\), which can also take the deterministic form \(a_t=\mu_\theta(s_t)\). The next state \(s_{t+1}\) is then generated by the dynamics of the environment, that are restricted to satisfy the markov property \(p(s_{t+1}|s_t,a_t)=p(s_{t+1}|s_1,a_1...s_t,a_t)\). The trajectories are thus sampled from a markov decision process (MDP). The state \(s_t\) sometimes can be only partially observed (\cite{hausknecht2015deep}), which yields a partially observable markov decision process (POMDP).

\chapter{Variants of Reinforcement Learning}
\label{ch:variantsOfRL}

The goal of episodic reinforcement learning algorithms is to maximize the expected accumulated reward or return \(R_1=\mathbb{E}_{\pi}(\sum_{i=1}^{T-1}{\gamma^{i-1}r_{i+1}})\) for all initial states \(s_1\), where T denotes the end of the episode, \(r_{t+1}\) the reward for state \(s_t\) and action \(a_t\), and \(\gamma\in[0,1]\) is a discounting factor for future rewards. At each time step, the direct reward depends on the current state and the action chosen by the model and thus on the executed policy. Theoretically, many sampled trajectories including all possible states and actions at different positions in time would be needed to infer a policy that reliably maximizes the expected return over all possible trajectories. For continuing environments, an estimate of the average reward per time step denoted as \(\sum_{s\in S} p_{\pi}(s)\sum_{a\in A} \pi(a|s)r(s,a)\) can be used as a measure of the policy quality, where \(S\) and \(A\) are the respective sets of all states and actions and \(p_{\pi}(s)\) is the probability of arriving at state \(s\) when following policy \(\pi\) (\cite{sutton1998reinforcement}). While we focus on episodic reinforcement learning in the following, it is possible to apply most of the discussed algorithms also to continuing environments.

\section{Common Reinforcement Learning Issues}
\label{sec:rlissues}

Many recent improvements in the field of reinforcement learning are caused by the difficulties that occur, when trying to apply standardized algorithms in realistic environments. In the following, we will first name some of the main difficulties reinforcement learning algorithms are confronted with and then present various methods researchers have invented to overcome these limitations, starting with relatively simple tabular methods that have been used for a long time and continuing with recent approaches to improve training performance and especially enable algorithms to deal with highly complex environments.

\textbf{Exploring the state space:}\\
In contrast to randomly exploring the state space, especially deterministic policies, when they are used to sample trajectories, do not visit large parts of the state space and thus lead to algorithms converging to suboptimal solutions. During sampling it is required to deliberately include suboptimal actions to explore unseen parts of the state space. Whether to greedily choose the assumed best action or a random action is known as the exploitation-exploration tradeoff (\cite{kaelbling1996reinforcement, woergoetter2008reinforcement}).

\textbf{Complex state spaces:}\\
In practice, reinforcement learning algorithms should be able to deal with large state spaces like pixel images that could be observations of a camera (\cite{mnih2013playing}). This imposes the difficulty of learning a policy even when only a small subset of all possible states will be visited during training. The fact that any image generated by a random generator almost never looks like any realistic scene, although there is a non-zero probability for it to do so, suggests that comparatively few examples of an image of given size correspond to natural images (\cite{goodfellow2016deep}). Furthermore, some of the possible states might not be reached during training due to restricted training time. Images from cameras or other complex state representations might also appear noisy, which again complicates the task of learning a good policy.

\textbf{Continuous action spaces:}\\
For realistic environments it is often not sufficient to design a policy that deterministically outputs a discrete action or stochastically models a probability distribution over a set of discrete actions. In robotic applications, motor torques must be produced that can take any continuous value and must be exact to make the robot move correctly. Although it is sometimes possible to successfully discretize a continuous action space, using this approach always means losing flexibility (\cite{lillicrap2015continuous}). For problems with multiple continuous actions, like producing multiple motor torques in parallel, the number of corresponding discrete actions rises exponentially, which quickly makes following this solution impossible.

\textbf{High variance of the trained estimator:}\\
While the policy estimation should normally converge to an optimum when the number of training examples is large, practical environments induce strong correlations between samples that are temporally close to each other. A robot might be expected to behave very similarly in close situations and the state representation might not change significantly. Thus, the variance of the trained estimator will be high, as the samples from the environment change slowly and the learned model will always tend to overfit the training data currently presented and forget important past transitions (\cite{lin1992self}). This means it will more likely fail to generalize to other data. The lack of generalization will at least negatively affect the training performance or the algorithm will not be able to learn a resonable policy at all.

\textbf{Partially observable environments:}\\
For some environments it might be impossible to observe the complete state \(s_t\). A robotic camera for instance might not be able to capture the whole scene with all objects relevant for the task, but only parts of it. The impact of the action that has to be generated by the policy depends on the system dynamics, which base on \(s_t\). That makes it necessary to gather more information by memorizing multiple observations to predict a reasonable action (\cite{hausknecht2015deep}).

\section{Algorithms that Optimize Value Functions}
\label{sec:valueFunctionAlgorithms}

\subsection{Reinforcement Learning with Tabular Value Functions}
\label{sec:rlWithTabularValueFunctions}

The expected return for the initial state with respect to the policy is the target for optimization, like discussed above, as it spans the whole trajectory. However, only estimating \(R_1\) is not sufficient to derive an optimal policy, because the policy also needs to output an optimal action for every intermediate state. If both the state and action space are discrete, the straightforward way to find an optimal policy and thus maximize the expected return over trajectories is to either estimate the optimal state-value function \(V^*\) or the optimal action-value function \(Q^*\) (\cite{sutton1998reinforcement}). The state-value function specifies the expected return of a state \(s_t\) when following an optimal policy, whereas the action-value function specifies the expected return when choosing action \(a_t\) in state \(s_t\) and following an optimal policy subsequently.

\begin{equation}
V^{*}(s_t)= \max_{a_t}{\mathbb{E}[r_{t+1}+\gamma V^*(s_{t+1})]}
\end{equation}
\begin{equation}
Q^{*}(s_{t},a_{t})= \mathbb{E}[r_{t+1}+\gamma\max_{a_{t+1}}{Q^*(s_{t+1},a_{t+1})}]
\end{equation}

With good estimates of all \(V^*(s_t)\), an optimal greedy policy can be easily derived by always choosing the action which most probably leads to the best rated states. To estimate the states the environment might take after executing an action, it is required to know the dynamics of the system. For system dynamics \(p(s_{t+1}|s_t,a_t)\) action \(a_t\) would be chosen according to:

\begin{equation}
\argmax_{a_t}{\sum_{s_{t+1}} p(s_{t+1}|s_t,a_t)V^*(s_{t+1})}
\end{equation}

With good estimates of all \(Q^*(s_t,a_t)\) it is no longer necessary to know the dynamics of the system, because the action-value function implicitly captures the transition probabilities. Many learning algorithms thus focus on approximating \(Q^*\) rather than \(V^*\). These algorithms can be applied to a broad range of environments with different dynamics without the need to train a model of the environment and are thus called \textit{model-free} (\cite{kaelbling1996reinforcement}). The optimal action with respect to \(Q^*\) can be chosen as simply as:

\begin{equation}
\argmax_{a_t}Q^*(s_t,a_t)
\end{equation}

To form a learning algorithm, the action-value function is redefined with respect to an arbitrary policy \(\pi\). The value of \(Q^\pi(s_t,a_t)\) corresponds to the expected return when taking action \(a_t\) in step \(s_t\) and following policy \(\pi\) subsequently. The calculated estimates can then be used to improve the policy. Intuitively, for optimal \(\pi\), the action-value function with respect to the policy converges to \(Q^{*}(s_t,a_t)\). It is possible to estimate \(Q^\pi(s_t,a_t)\) for example using a Monte-Carlo estimate of the expected return after the end of each episode. This strategy is called \textit{offline} learning, because it must wait until an episode has finished. However, it is much more practical to store estimates of \(Q^\pi(s_t,a_t)\) in an array \(Q(s_t,a_t)\) and reuse existing estimates for future states, which is called \textit{bootstrapping} and yields an \textit{online} learning rule (\cite{sutton1998reinforcement}):

\begin{equation}
Q(s_t,a_t) \pluseq \alpha(r_{t+1}+\gamma Q(s_{t+1},a_{t+1})-Q(s_{t},a_{t}))
\end{equation}

The transitions between states are sampled from the system dynamics and the actions from the policy \(\pi\). Continuously updating the estimates of \(Q^\pi(s_t,a_t)\) while improving the policy to follow the maximal Q-values leads to the SARSA algorithm presented in Listing \ref{lst:sarsa}.

\begin{minipage}{\linewidth}
\begin{lstlisting}[language=Matlab,mathescape,label={lst:sarsa}, caption={SARSA (State-Action-Reward-State-Action), reproduced in essence from \textcite{sutton1998reinforcement}. }, captionpos=t,backgroundcolor=\color{white},frame=lines]
Initialize Q(s,a) arbitrarily
Repeat (for each episode):
    Initialize $s_1$
    Choose $a_1$ from $s_1$ using policy derived from $Q$
    Repeat (for each $t$ of episode):
        Take action $a_t$, observe $r_t$, $s_{t+1}$
        Choose $a_{t+1}$ from $s_{t+1}$ using policy derived from $Q$
        $Q(s,a)\leftarrow Q(s,a)+\alpha[r_{t+1}+\gamma Q(s_{t+1},a_{t+1})-Q(s_t,a_t)]$
        $s_t\leftarrow s_{t+1}; a_t\leftarrow a_{t+1};$
\end{lstlisting}
\end{minipage}

SARSA performs \textit{on-policy} bootstrapping, because the expected return \(Q(s_{t+1},a_{t+1})\) of the next state depends on the policy \(\pi\). It is also possible to alter the update rule to bootstrap using the best action in the next state yielding the popular Q-learning algorithm introduced by \textcite{watkins1992q}, which is presented in Listing \ref{lst:Qlearning}. This algorithm can be viewed as directly approximizing the action-value function of an optimal policy, as the policy used for bootstrapping is directly given by the algorithm and does not necessarily match the policy \(\pi\) used for sampling. Therefore Q-learning is called an \textit{off-policy} algorithm. The Q-values in Q-learning are updated as follows:

\begin{equation}
Q(s_t,a_t) \pluseq \alpha(r_{t+1}+\gamma max_{a}Q(s_{t+1},a)-Q(s_{t},a_{t}))
\end{equation}

\begin{minipage}{\linewidth}
\begin{lstlisting}[language=Matlab,mathescape,label={lst:Qlearning}, caption={Q-Learning, reproduced in essence from \textcite{sutton1998reinforcement}. }, captionpos=t,backgroundcolor=\color{white},frame=lines]
Initialize Q(s,a) arbitrarily
Repeat (for each episode):
    Initialize $s_1$
    Repeat (for each $t$ of episode):
        Choose $a_{t}$ from $s_{t}$ using policy derived from $Q$
        Take action $a_t$, observe $r_t$, $s_{t+1}$
        $Q(s,a)\leftarrow Q(s,a)+\alpha[r_{t+1}+\gamma \max_{a}Q(s_{t+1},a)-Q(s_t,a_t)]$
        $s_t\leftarrow s_{t+1};$
\end{lstlisting}
\end{minipage}

\subsection{Using Stochastic Policies to Improve Exploration}
\label{sec:stochasticPolicyExploration}

One beneficial property of an off-policy learning approach like Q-learning (Listing \ref{lst:Qlearning}) is that any policy can be used for sampling, while the learning rule is not affected. It is in particular possible to design a policy for sampling that encourages exploration and thus leads to a better search in the state space as stated above, while the learned policy is still greedy.

An exploring policy needs to be based on some kind of randomization to detect unknown or less frequently visited regions in state space. Policies of that kind do not deterministically predict an action to execute but rather model a probability distribution from which the actions will be sampled. With off-policy training algorithms it is possible to train a deterministic policy while using a stochastic policy for sampling.

A popular choice for a policy that can be used to encourage exploration during sampling, when in most cases the best action known so far (greedy action) should be executed is the \textbf{\(\epsilon\)-greedy} policy (\cite{sutton1998reinforcement}). The parameter \(\epsilon\in[0,1]\) specifies the probability of selecting a completely random action. In all other cases the assumed best action will be executed:

\begin{equation}
a_t =
\begin{cases}
    \text{a random action} & \text{with probability} \, \epsilon\\
    \argmax_a Q(s_t, a) & \text{otherwise}
\end{cases}
\end{equation}

This strategy however ignores the fact, that there might be large differences between the Q-values of the non-optimal actions, which are all chosen with equal probability. The \textbf{softmax} policy (\cite{sutton1998reinforcement}) consideres this fact and assigns a different probability to each action that depends on the Q-value of the action:

\begin{equation}
p(a_t)_{s_t}=\frac{e^{Q(s_t,a_t)/\tau}}{\sum_{a}e^{Q(s_t,a)/\tau}}
\end{equation}

The parameter \(\tau\) is called the temperature and regularizes the amount of randomization induced. When \(\tau\) decreases, the policy becomes more and more deterministic or greedy, while increasing \(\tau\) encourages exploration.

\subsection{Using Eligibility Traces to Improve Bootstrapping}
\label{sec:eligibilityTraces}

The bootstrapping idea presented in section \ref{sec:rlWithTabularValueFunctions} only supports direct updates of the value function estimates with respect to the estimated value of the following action. This can considerably slow down learning, especially if there are long chains of consecutive actions. A robot might need to appoach a target for a long time before reaching it and thus might get a reward only after executing many steps beforehand. In this case, the reward associated with the last action is known as a \textit{delayed reward} (\cite{watkins1989learning}). To obtain positive value estimates for the first actions in the chain, many episodes are needed, because during the first episode only the value of the last action is updated, then the value of the second last action due to bootstrapping and so on. This problem is also known as the credit assignment problem (\cite{woergoetter2008reinforcement}), as present actions might be essential for future rewards, but do not directly benefit from it.

With Monte Carlo estimates it is not necessary to repeatedly visit the same consecutive actions multiple times to obtain valid value estimates at the beginning of the chain, because the accumulated return over all future actions is directly used to form the estimates. Using Monte Carlo methods however causes other issues like high variance. It is possible to overcome this tradeoff by using \textit{n-step} bootstrapping, that spans over multiple steps and can also include weights to raise the impact of rewards to temporally close actions (\cite{sutton1998reinforcement}). N-step bootstrapping provides an intermediate solution that combines the advantages of Monte Carlo estimates and direct bootstrapping from the next action-value.

The Q-Learning algorithm can be viewed as comparing the expected return by choosing a specific action to the current action-value estimate. The expected return is obtained using bootstrapping from the next action value, called \textit{one-step} bootstrapping. The update of the current Q-value with a learning rate \(\alpha\) is then given by:

\begin{align}
\Delta Q(s_t,a_t)=\alpha(R_t^{(1)}-Q(s_t,a_t))\\
R_t^{(1)}=r_{t+1}+\gamma max_{a}Q(s_{t+1},a)
\end{align}

The term \(R_t^{(1)}\) is called the \textit{one-step} return. In contrast, the Monte Carlo estimate starting from state t would accumulate the rewards till the end of the episode without any bootstrapping:

\begin{equation}
\label{eq:monteCarloReturn}
R_t=\sum_{i=t}^{T-1}{\gamma^{i-t-1}r_{i+1}}
\end{equation}

The n-step return mixes both views and bootstraps from the value of the action executed n timesteps after t. If the episode ends earlier, the n-step return is equivalent to the Monte Carlo estimate \(R_t\).

\begin{equation}
R_t^{(n)}=\gamma^{n} max_{a}Q(s_{t+n},a)+\sum_{i=t}^{t+n-1}{\gamma^{i-t-1}r_{i+1}}
\end{equation}

As the best value of n is hard to predict, but it is straightforward to see that temporally closer actions should have a greater effect on each other, we can accumulate all n-step returns with a decaying weighting factor \(\lambda^{(n-1)}\):

\begin{equation}
R_t^{(\lambda)}=(1-\lambda)\sum_{n=1}^{\infty}\lambda^{n-1}R_t^{(n)}
\end{equation}

It is even possible to formulate an online learning algorithm using this idea, by memorizing actions executed in the past. The \textbf{eligibility trace} (\cite{sutton1988learning}) assigns a value to each pair of state and action that is reset to 1, when the action is executed in the respective state and then decays by \(\gamma\lambda\) after every time step. Q-Learning with eligibility traces is called Q(\(\lambda\)) learning and updates all Q-values at every time step as follows:

\begin{align}
\forall_{s,a}: Q(s,a)=Q(s,a)+\alpha\delta_t e(s,a)\\
\delta_t=r_{t+1}+\gamma \max_{a} Q(s_{t+1},a)-Q(s_t,a_t)\\
\forall_{s,a}: e(s,a)=
\begin{cases}
1 &if(s,a)=(s_t,a_t)\\
\gamma\lambda e(s,a) &otherwise
\end{cases}
\label{eq:replacingEligibilityTrace}
\end{align}

\begin{figure}[ht]
\centering
\includegraphics[width=0.78\textwidth]{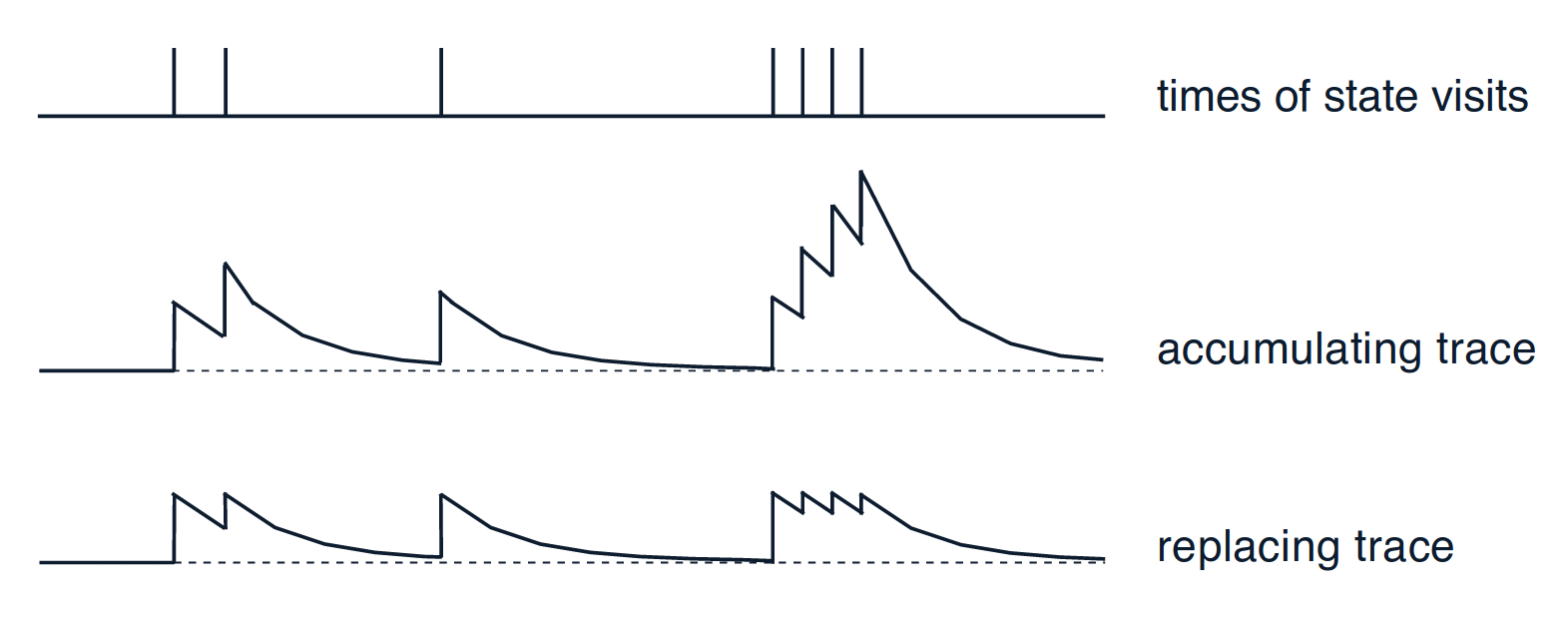}
\caption{Accumulating eligibility traces do not have a fixed maximum value, but can grow larger than 1, when a pair of state and action is visited often. Figure adapted from \textcite{sutton1998reinforcement}. }
\label{fig:eligibilityTraces}
\end{figure}

The term \(\delta_t\) is known as the \textit{TD-error} for step t and \(e(s,a)\) is the eligibility trace assigned to state s and action a. It is straightforward to see, that eligibility traces mix the ideas of Monte Carlo estimates and one-step bootstrapping, as the update rules are equivalent to Monte Carlo estimates for \(\lambda=1\) and to one-step bootstrapping for \(\lambda=0\). In practice, further adjustments might be necessary to regard the case, when exploration happens (see section \ref{sec:stochasticPolicyExploration}) and thus the chain of assumed optimal actions is broken (\cite{watkins1989learning}). It is also possible to accumulate the value of eligibility traces, when a pair of state and action is visited often in a short length of time. Figure \ref{fig:eligibilityTraces} compares replacing traces used in equation \ref{eq:replacingEligibilityTrace} to accumulating traces.

\subsection{Using Neural Networks to Approximate the Q-function}

The tabular methods described in the sections above do not apply to large discrete or continuous state spaces, which require a specific function approximator to estimate the Q-function. Estimating the Q-function can be done by a supervised learning algorithm with the targets for training given by the reinforcement learning algorithm. Therefore a loss function is introduced that drives the function approximator to output the correct Q-values, where \(Q(s_{t},a; \bm{\theta})\) is a function, parametrized by learned parameters \(\bm{\theta}\):

\begin{equation} \label{eq:dqnSimpleLoss}
\mathcal{L}(s_t,a_t,r_{t+1},s_{t+1},\bm{\theta})=(r_{t+1}+\gamma max_{a}Q(s_{t+1},a; \bm{\theta})-Q(s_t, a_t; \bm{\theta}))^2
\end{equation}

\begin{figure}[ht]
\centering
\includegraphics[trim={0 0.8cm 0 0}, clip,width=0.78\textwidth]{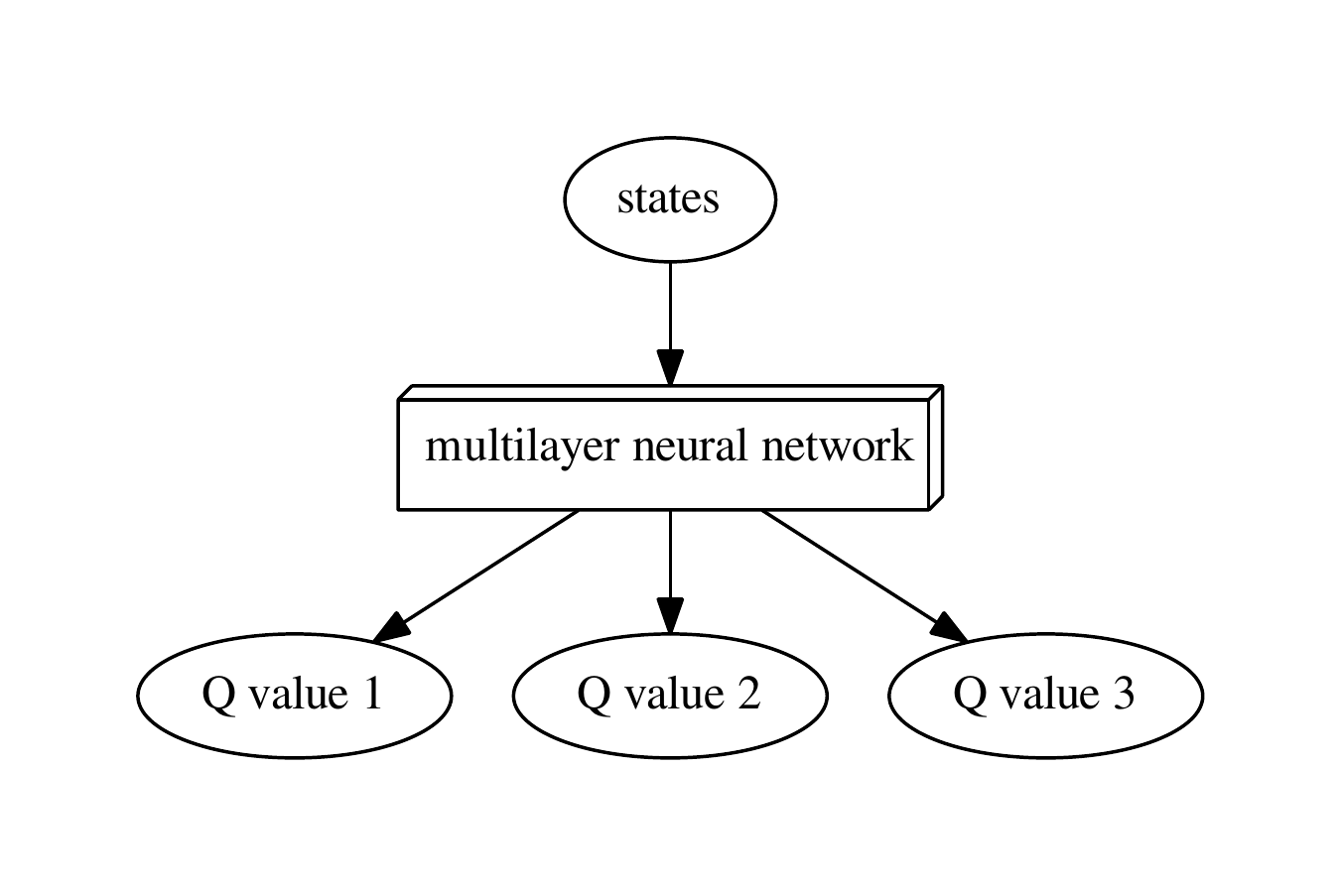}
\caption{Architecture of a simple deep Q-network with 3 discrete actions. }
\label{fig:deepQ}
\end{figure}

Neural networks have proven to be effective for function approximation and supervised learning tasks. Architectures solving supervised learning tasks, like estimating the Q-values in the reinforcement learning setting, usually take the form of feed-forward neural networks. This means, the neurons are organized in layers, that are ordered and connected in only one direction from the input layer to the output layer. Normally, there are no connections between neurons in a single layer. Deep feed-forward neural networks with multiple intermediate layers (hidden layers) between input and output layer are able to deal with complex state spaces and can generalize to unknown states that were never observed during training (\cite{levine2016end}). A neural network with only feed-forward connections, non-linear activation functions and one or more hidden layers is also known as a multi-layer perceptron (MLP). The architecture of a simple Q-network is depicted in figure \ref{fig:deepQ}. The network takes the state representation as input and outputs the Q-values for all actions using seperate output neurons. The action space thus still needs to be discrete to enable the network to learn the Q-values for all actions.

\subsection{Using Convolutional Layers to Process Pixel Images}

To extract useful information from pixel images, it is often not sufficient to only use fully connected feed-forward layers, where every neuron of one layer is connected to every unit of the next layer. Like required by other supervised learning tasks, in the reinforcement learning setup, neural networks should be able to perform a generalization task and reduce the complexity of the state space to being able to correctly predict the Q-values later. While it is theoretically possible to capture the information presented in an image by purely using fully connected layers, network structures like these induce several problems.

First of all, a great number of neurons and thus even more connections would be needed, because each pixel in an image must be represented by a single neuron or even multiple neurons for multiple color channels. To store the associated parameters, massive amounts of memory would be required and it would also take comparatively long time to calculate the outputs of the network for given inputs. Because of the large amount of parameters, fully connected layers converge slowly and can even completely fail to capture the relevant information of larger pixel images, if not carefully designed. Downsampling of the images reduces the complexity of the network, but also discards a lot information.

\begin{figure}
\centering
\includegraphics[width=0.78\textwidth]{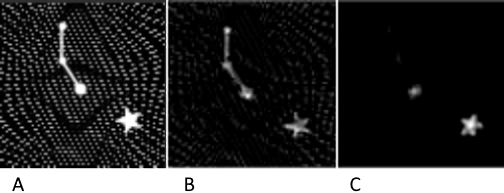}
\caption{Convolutional layers can be used to detect features in images and effectively eliminate background noise. \textit{(a)} Sample activations of the input layer of a neural network. \textit{(b)} Corresponding activations of neurons in the first hidden layer, produced by a filter that detects mainly low level features like edges. \textit{(c)} Corresponding activations of neurons in the second hidden layer, produced by a filter that detects high level features like the position of the star in the image.}
\label{fig:convolution}
\end{figure}

Convolutional networks (CNNs) were successfully applied by \textcite{krizhevsky2012imagenet} to win the ImageNet challenge with an exceptional good score. The convolution operation applied to neural networks deliberately restricts the connections between two layers to be local. This is a reasonable assumption, as the pixels in an image, that are close to each other, are normally much stronger correlated. Each unit of the first hidden layer is thus a function of only a small patch of the input image. Furthermore, the connections between each image patch and the corresponding unit in the next layer are restricted to be equal over the whole image, which again reduces the amount of parameters needed. Figure \ref{fig:convolution} shows a simple example of convolutions applied to identify an object in an image.

The filter function that slides over the image to produce the next neural layer is called a \textit{kernel} and output of the convolution operation a \textit{feature map} of the input (\cite{goodfellow2016deep}). Most of the time, an additional activation function is executed after the convolution operation to transform the elements of the feature map to the activations of the next layer. Multiple learned kernels can be used to produce multiple feature maps for detecting different features in the image. Stacking convolutional layers enables the network to learn a hierarchical form of dependencies between pixels in distant regions of the original input image.

Especially for applications that are only interested in extracting features from images and not in where these features occur in the image, it is useful to downsample the intermediate feature maps. This operation is called \textit{pooling} (\cite{scherer2010evaluation}). Fully connected layers can then be used to further process the output after several convolutional layers, which has greatly reduced dimensionality. Convolutional layers can also be transposed to reconstruct image data from a lower dimensional representation.

Each activation in a feature map produced by a convolutional layer is a function of the convolution kernel and the input. For X being the two-dimensional input image, K the kernel and Y the elements of the feature map, the convolution operation (denoted as \(*\)) can be applied to transform X to Y:

\begin{equation}
\label{eq:convOperation}
Y=X*K
\end{equation}

The convolution operation flips the kernel to obtain a commutative operation. If a two-dimensional matrix kernel is used, that has odd height m and width n to be centered on one pixel whose coordinates in the kernel are defined to be (0,0), equation \ref{eq:convOperation} decomposes to:

\begin{equation}
Y(i,j)=(X*K)(i,j)=\sum_{s_1=\frac{-m+1}{2}}^{\frac{m-1}{2}}\sum_{s_2=\frac{-n+1}{2}}^{\frac{n-1}{2}}X(i-s_1,j-s_2)K(s_1,s_2)
\end{equation}

\subsection{Deep Q-network (DQN)}
\label{sec:dqn}

\textcite{mnih2013playing} showed that deep learning with convolutional layers can enable reinforcement learning algorithms to successfully learn to play Atari 2600 games. An improved version of this approach was presented later as deep Q-network (\cite{mnih2015human}), that was able to use direct training from pixels to actions to play 49 different Atari games without the need to change the hyperparameters of the network or make any other modifications for a specific game. \textcite{zhang2015towards} showed how to train a robotic arm with the DQN approach to perform reaching tasks while only observing camera images. The performance on Atari games is comparatively impressive, as the learned policies were often able to outperform human players and only the pixel images and the game score were used as input to the training algorithm, which means that there was no domain knowledge available to the algorithm.

While some Atari games can be directly modeled as fully observable MDPs as discussed in section \ref{sec:rl}, it is not possible to infer properties like the velocity of objects from a single image, which is necessary to establish effective strategies for some games. Therefore, a sequence of four frames is passed into the network to provide the missing information to the network and approximately satisfy the markov property.

The success of DQN however does not only rely on the usage of a neural network function approximator. There are some problems that would practically prevent neural networks as nonlinear function approximators from converging. First, the loss function given in equation \ref{eq:dqnSimpleLoss} includes the parameters \(\bm{\theta}\) twice, which arguably makes learning instable. The foresight into the future \(Q(s_{t+1},a; \bm{\theta})\) to be used for bootstrapping should not directly depend on \(\bm{\theta}\) to stabilize learning and reduce the variance of the approximated function. The DQN training method therefore introduces a \textbf{target Q-network}, that copies the parameters from the trained Q-network only after several hundred or thousand training steps and thus does not change rapidly and enables the algorithm to learn stable long term dependencies (\cite{mnih2015human}). With parameters \(\bm{\theta^{-}}\) of the target network, the loss function changes to:

\begin{equation}
\label{eq:dqnLossWithTargetNetworks}
\mathcal{L}_{DQN}(s_t,a_t,r_{t+1},s_{t+1},\bm{\theta},\bm{\theta}^{-})=(r_{t+1}+\gamma max_{a}Q(s_{t+1},a; \bm{\theta^{-}})-Q(s_t, a_t; \bm{\theta}))^2
\end{equation}

Unfortunately, simple gradient descent on the loss function with target network can still lead to high variance of the function estimator due to the inherent structure of the learned data. Especially if collecting more training examples is costly as it is with data generated by a real robotic system, other ways need to be found to encourage generalization of the trained network. The technique used for training the DQN is essentially equivalent to \textit{stochastic gradient descent} on a memory of past transitions of the reinforcement learning environment called the \textbf{experience replay} memory (\cite{lin1992self}). The idea behind stochastic gradient descent is to use random samples of relatively few training examples to estimate the expectation of the true training error. When the examples are sampled from very different time steps and were generated under different conditions, they can be sufficient to provide a
 good estimate of the true trainig error with relatively low variance. The experience replay memory stores transitions between states sampled in the past, even for multiple episodes, and also memorizes the corresponding actions and rewards to being able to correctly calculate the loss at every time step in the future. Thus, the memory consists of samples \((s_i,a_i,r_{i+1},s_{i+1})\) for each recorded time step.

\begin{minipage}{\linewidth}
\begin{lstlisting}[language=Matlab,mathescape,label={lst:dqn}, caption={Deep Q-Network (DQN) with experience replay and target network, adapted from \textcite{mnih2015human}.}, captionpos=t,backgroundcolor=\color{white},frame=lines]
Initialize replay memory $D$
Initialize action-value function $Q$ with random weights $\bm{\theta}$
Initialize target action-value function $\hat{Q}$ with weights $\bm{\theta}^{-} = \bm{\theta}$
for episode = 1 to $M$ do
    Initialize sequence $s_1 = \{x_1\}$ and preprocessed sequence $\phi_1 = \phi(s_1)$
    for $t$ = 1 to $T$ do
        Select $a_t$ = $\begin{cases}
                \text{a random action} & \text{with probability} \, \epsilon\\
                \argmax_a Q(\phi(s_t), a; \bm{\theta}) & \text{otherwise}
        \end{cases}$

        Execute action $a_i$ in emulator and observe reward $r_t$ and image $x_{t+1}$
        Set $s_{t+1} = s_t$, $a_t$, $x_{t+1}$ and preprocess $\phi_{t+1} = \phi(s_{t+1})$
        Store transition $(\phi_t , a_t , r_t , \phi_{t+1})$ in $D$

        // sample from experience replay memory
        Sample random minibatch of transitions $(\phi_j , a_j , r_j , \phi_{j+1})$ from $D$
        Set $y_j$ = $\begin{cases}
                r_j & \text{if episode terminates at step} \, j + 1 \\
                j + \gamma \max_{a'} \hat{Q}(\phi_{j+1}, a';\bm{\theta}^{-}) & \text{otherwise}
        \end{cases}$
        Perform a gradient descent step on $(y_j-Q(\phi_j, a_j; \bm{\theta}))^2$ w.r.t. to the network parameters $\bm{\theta}$

        // update target network
        Every $C$ steps reset $\hat{Q}$ = $Q$, that means, set $\bm{\theta}^{-} = \bm{\theta}$
    end for
end for
\end{lstlisting}
\end{minipage}

Let \(m\) denote the number of all examples stored in the experience replay memory and \(m'\) a relatively small number of examples, which are supposed to be randomly sampled from the memory, called the \textit{batch size}, where \(m'\ll m\). The goal is to minimize the loss for all examples \(J(\bm{\theta})=\frac{1}{m}\sum_{i=1}^{m}L(s_i,a_i,r_{i+1},s_{i+1},\bm{\theta})\). For simplicity, the uninteresting dependency of the loss function on \(\bm{\theta}^{-}\) is left out here, but could be easily added. The sum of all losses can be approximized as follows:

\begin{equation}
J(\bm{\theta})=\frac{1}{m}\sum_{i=1}^{m}\mathcal{L}(s_i,a_i,r_{i+1},s_{i+1},\bm{\theta})\approx\frac{1}{m'}\sum_{i=1}^{m'}\mathcal{L}(s_i,a_i,r_{i+1},s_{i+1},\bm{\theta})
\end{equation}

Gradient descent can then be intuitively performed by shifting the parameters in the direction of the negative gradient:

\begin{equation}
\bm{g}=\nabla_{\bm{\theta}}J(\bm{\theta})\approx\frac{1}{m'}\nabla_{\bm{\theta}}\sum_{i=1}^{m'}\mathcal{L}(s_i,a_i,r_{i+1},s_{i+1},\bm{\theta})
\end{equation}

\begin{equation}
\label{eq:parameterUpdate}
\bm{\theta}\minuseq\alpha\bm{g}
\end{equation}

Listing \ref{lst:dqn} shows the DQN algorithm of the original publication in pseudocode including all main concepts discussed so far, namely: using neural networks as function approximators for the action-value function (Q-function), including target networks for bootstrapping of the action-value function, and using experience replay as a variant of stochastic gradient descent.

The parameter \(\alpha\) in equation \ref{eq:parameterUpdate} denotes the learning rate. In fact, simple gradient descent like this can work reasonably well for a learning rate that is appropriate for the optimized problem, but also very likely fails to converge for a randomly chosen and fixed learning rate. While it is possible to find a suitable learning rate by searching the the hyperparameter space, this would be computationally expensive as many training steps need be done to evaluate a single learning rate. Gradient descent can be seen as navigating through the hyperplane spanned by the unified loss \(J(\bm{\theta})\) (error function) and all parameters \(\bm{\theta}_i\) being the coordinates. Learning with any fixed learning rate thus can be slow, because it takes long time to move across flat regions, where the gradient is very small. When gradient descent arrives at a specific point of the hyperplane, it is possible for the error function to rapidly decrease in only one or few directions of the parameter space while moving in other directions leaves the error nearly unchanged (\cite{goodfellow2016deep}). This motivates the use of learning algorithms that adapt to the shape of the parameter space.

\begin{figure}[ht]
\centering
\includegraphics[width=0.5\textwidth]{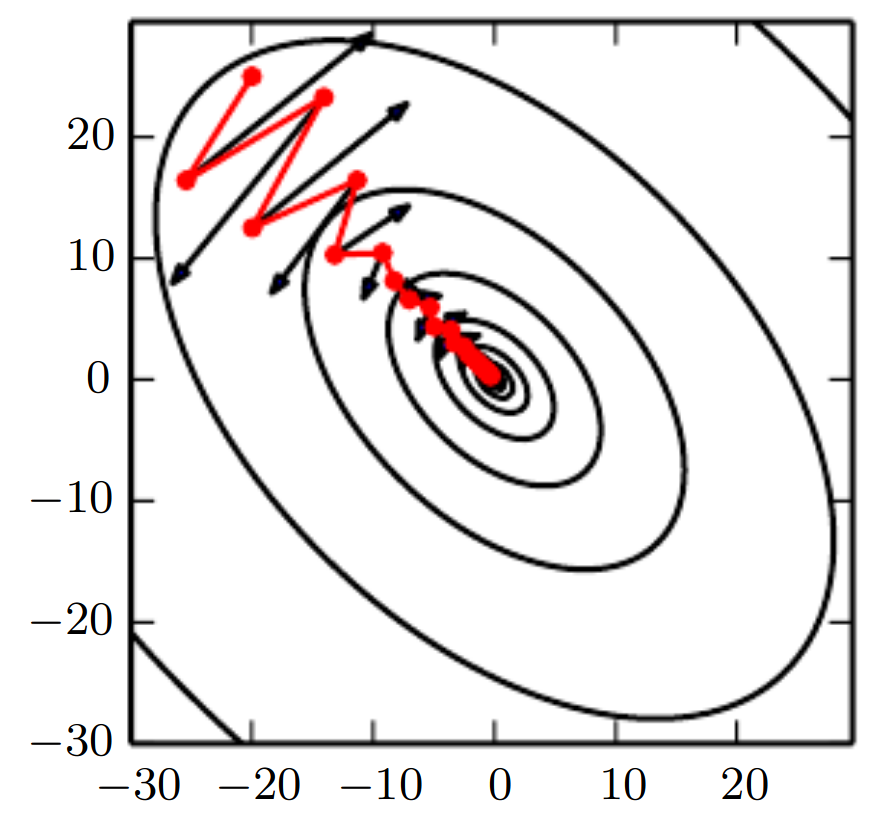}
\caption{With the additional momentum term, gradient descent arrives faster at the minimum of the cost function without wasting too much time for oscillation. Figure adapted from \textcite{goodfellow2016deep}. }
\label{fig:momentum}
\end{figure}

The RMSProp algorithm introduced by \textcite{tieleman2012lecture} is able to sidestep the issue of a fixed learning rate by using a seperate learning rate for each direction of the parameter space and automatically adapting these learning rates to the magnitude of the gradient for the respective direction. It is often beneficial to include an additional momentum term in the parameter update that plays the same role like velocity in physics. This term is for example good for reducing the probability of oscillation, when the hyperplane for which the minimum should be found looks like a valley with steep sides. An example of such a situation is depicted in figure \ref{fig:momentum}. The Adam algorithm (\cite{kingma2014adam}) is a modification of RMSProp, that includes a momentum term by default.

\subsection{Improvements to DQN}
\label{sec:dqnImprovements}
\textbf{Double DQN (D-DQN)} as proposed by \textcite{van2016deep} was again able to improve the performance of DQN applied to Atari games by a minor modification of the training target. The loss function with target network given in equation \ref{eq:dqnLossWithTargetNetworks} is considered problematic, because it tends to overestimate the future action values. Both the selection and evaluation of future actions depend on the parameters of the target network \(\bm{\theta}^{-}\). Equation \ref{eq:dqnLossWithTargetNetworks} can thus be rewritten as:

\begin{align*}
\mathcal{L}_{DQN}&(s_t,a_t,r_{t+1},s_{t+1},\bm{\theta},\bm{\theta}^{-})=\\
&(r_{t+1}+\gamma Q(s_{t+1},\argmax_{a}Q(s_{t+1},a; \bm{\theta}^{-}); \bm{\theta}^{-})-Q(s_t, a_t; \bm{\theta}))^2 \numberthis
\end{align*}

The loss function used by D-DQN disentangles the action selection from the evaluation of the selected action by using the trained parameters \(\bm{\theta}\) to select future actions instead of those of the target network:

\begin{align*}
\label{eq:lossDDQN}
\mathcal{L}_{D-DQN}&(s_t,a_t,r_{t+1},s_{t+1},\bm{\theta},\bm{\theta}^{-})=\\
&(r_{t+1}+\gamma Q(s_{t+1},\argmax_{a}Q(s_{t+1},a; \bm{\theta}); \bm{\theta}^{-})-Q(s_t, a_t; \bm{\theta}))^2 \numberthis
\end{align*}

Another popular approach to improve DQN called a \textbf{dueling network architecture} is proposed by \textcite{wang2015dueling}. The authors do not directly use the neural network to predict the Q-function. The network itself predicts two functions of the input with respect to the policy currently followed: A state-value function \(V^{\pi}(s_t)\) and an advantage function \(A^{\pi}(s_t,a_t)\) defined as:

\begin{equation}
\label{eq:advantageFunction}
A^{\pi}(s_t,a_t)=Q^{\pi}(s_t,a_t)-V^{\pi}(s_t)
\end{equation}

Both outputs are unified to form an estimate of the Q-value which is then used for training. The loss to be optimized can therefore include all previously made improvements and for instance take the form of equation \ref{eq:lossDDQN}. It is also possible to use experience replay and a target network exactly like discussed above. The straightforward way to derive \(Q^{\pi}(s_t,a_t)\) from \(V^{\pi}(s_t)\) and \(A^{\pi}(s_t,a_t)\) would be to simply reorganize equation \ref{eq:advantageFunction}:

\begin{equation}
\label{eq:advantageFunctionReorganized}
Q^{\pi}(s_t,a_t)=A^{\pi}(s_t,a_t)+V^{\pi}(s_t)
\end{equation}

As the learning algorithm only optimizes the Q-function and does not know anything about the underlying components, optimizing equation \ref{eq:advantageFunctionReorganized} almost never leads to \(V^{\pi}(s_t)\) or \(A^{\pi}(s_t,a_t)\) converging to good estimates of a state-value or advantage function. To force gradient descent on \(Q^{\pi}(s_t,a_t)\) to properly estimate these functions, a correction term is introduced:

\begin{equation}
Q^{\pi}(s_t,a_t)=A^{\pi}(s_t,a_t)-\max_{a}A^{\pi}(s_t,a)+V^{\pi}(s_t)
\end{equation}

When the assumed optimal action \(a^{*}=\argmax_{a}A^{\pi}(s_t,a)=\argmax_{a}Q^{\pi}(s_t,a)\) is chosen in state \(s_t\), all terms including the advantage function cancel out and the state-value function is optimized to be equal to the action-value function. This is a valid optimization, as the value of a state corresponds to the value of the best action to be chosen in that state, formally: \(V^{\pi}(s_t)=\max_{a}Q^{\pi}(s_t,a)=Q^{\pi}(s_t,a^{*})\). As \(V^{\pi}(s_t)\) is tied to be a valid estimate of the state-value function, the other terms naturally approximate the advantage function defined as the difference between state-value and action-value function. Other forms of the correction term like averaging over all possible actions are able to achieve a similar effect and yield better results in practice (\cite{wang2015dueling}).

\clearpage
\section{Algorithms that Follow the Policy Gradient}
Section \ref{sec:valueFunctionAlgorithms} showed that it is possible to derive reasonably performing policies from good estimates of value functions, especially useful are estimates of action-values provided by the Q-function. However, because policies derived from value functions search over a discrete number of Q-values to find the best action, it is not possible to directly obtain policies that output continuous actions using one of the methods described above.

Policy gradient methods directly parametrize the policy as a probability function \(\pi_{\bm{\theta}}(a_t|s_t)\) that can be completely described by the parameters \(\bm{\theta}\) and thus provide maximal freedom to learn any action-generating function. To evaluate different policies, the expected return following \(\pi\) over all trajectories conditioned by the policy, formally \(\tau\sim p_{\pi}(\tau)=p(\tau|\bm{\theta})\), is used. The return over a single trajectory \(r(\tau)\) is equal to the measure introduced at the beginning of chapter \ref{ch:variantsOfRL} and takes the form:

\begin{equation}
\label{eq:rewardSignal}
r(\tau)=\sum_{i=1}^{T-1}{\gamma^{i-1}r_{i+1}}
\end{equation}

The term \(r_{i+1}\) is the reward given to action \(a_i\) executed in state \(s_i\) of the respective trajectory. The probability distribution over trajectories \(p(\tau|\bm{\theta})\) decomposes as follows:

\begin{equation}
\label{eq:probTrajectory}
p(\tau|\bm{\theta})=p(\{s_1,a_1..s_T,a_T\}|\bm{\theta})=p(s_1)\prod_{i=1}^{T-1}\pi_{\bm{\theta}}(a_i|s_i)p(s_{i+1}|s_i,a_i)
\end{equation}

With these definitions, the target for optimization can be defined:

\begin{equation}
\label{eq:policyGradientTarget}
J(\bm{\theta})=\sum_{\tau\in T}p(\tau|\bm{\theta})r(\tau)
\end{equation}

While other methods like using evolutionary algorithms or random search are possible (\cite{deisenroth2013survey}), it is straightforward to optimize equation \ref{eq:policyGradientTarget} using gradient ascend:

\begin{equation}
\label{eq:updateThetaPolicyGradient}
\bm{\theta}\pluseq\alpha\nabla_{\bm{\theta}}J
\end{equation}

According to \textcite{peters2011policy}, following the policy gradient to solve reinforcement learning tasks only slightly modifies the parameters of the policy in contrast to value based methods, where large jumps between two estimated policies are possible. This property arguably improves training stability and convergence towards an optimal policy. In the following, several methods to estimate the gradient of the true expected return with respect to the parameters of the policy will be discussed.

\subsection{Finite-Difference Methods}
The main difficulty imposed by equation \ref{eq:updateThetaPolicyGradient} is to derive an appropriate estimate of \(\nabla_{\bm{\theta}}J\). Analytically calculating the gradient is impossible, as it would be necessary to sum over possibly infinitely many trajectories. The dynamics of the environment \(p(s_{t+1}|s_t,a_t)\) might also be unknown and not differentiable anyway. Finite-difference methods (FDM) are simple numerical methods to estimate first order gradients. FDM use the first two terms of the taylor series expansion and rearrange them to get an approximation of the true gradient. Among others, \textcite{peters2011policy} provide a compact description of how to apply FDM to find the policy gradient in the reinforcement learning setup. Let \(\bm{\theta}_{i}\) denote the i-th element of the parameter vector \(\bm{\theta}\) and \(\bm{\theta}+\alpha \bm{u}_i\) a small perturbation of the i-th component of \(\bm{\theta}\). The taylor series expansion approximates \(J(\bm{\theta}+\alpha \bm{u}_i)\) by using its partial derivatives:

\begin{equation}
J(\bm{\theta}+\alpha \bm{u}_i)
=J(\bm{\theta})
+\alpha\frac{\partial J(\bm{\theta})}{\partial \bm{\theta}_i}
+\frac{\alpha^2}{2}\frac{\partial^2 J(\bm{\theta})}{{\partial \bm{\theta}_i}^2}
+...
+\frac{\alpha^n}{n!}\frac{\partial^n J(\bm{\theta})}{{\partial \bm{\theta}_i}^n}
+R_n(\bm{\theta}+\alpha \bm{u}_i)
\end{equation}

The partial derivative of \(J\) with respect to \(\bm{\theta}_{i}\) can thus be approximated as follows:

\begin{equation}
\frac{\partial J(\bm{\theta})}{\partial \bm{\theta}_i}\approx\frac{J(\bm{\theta}+\alpha \bm{u}_i)-J(\bm{\theta})}{\alpha}
\end{equation}

To obtain the gradient for all components of the parameter vector, all partial derivatives must be approximated. The FDM approach has the beneficial property that it can be used for arbitrary policies, even for not differentiable policies. While the estimate of the gradient can be improved by averaging multiple estimated gradients for different perturbations, badly chosen perturbation can still make learning instable or cause it to fail (\cite{peters2011policy}). For realistic applications, the used policy can be assumed to be a differentiable function, which enables other estimations to work that are less error-prone and noisy.

\subsection{Likelihood-Ratio Methods}
\label{sec:likelihoodRatioMethods}

Let \(p(\tau|\bm{\theta})=p_{\bm{\theta}}(\tau)\) be the probability of trajectory \(\tau\) under policy \(\pi_{\bm{\theta}}\). The likelihood ratio trick, best known for its application in the REINFORCE algorithm introduced by \textcite{williams1992simple}, rewrites the gradient in equation \ref{eq:updateThetaPolicyGradient} using the property \(\nabla_{\bm{\theta}}\log p_{\bm{\theta}}(\tau)=\frac{\nabla_{\bm{\theta}} p_{\bm{\theta}}(\tau)}{p_{\bm{\theta}}(\tau)}\) as follows:

\begin{align}
\nabla_{\bm{\theta}}J(\bm{\theta})&=\sum_{\tau\in T}p_{\bm{\theta}}(\tau)\nabla_{\bm{\theta}}\log p_{\bm{\theta}}(\tau)r(\tau)\\
\label{eq:likelihoodRatioExpectation}&=\mathbb{E}_{\tau\sim p(\tau|\bm{\theta})}[\nabla_{\bm{\theta}}\log p_{\bm{\theta}}(\tau)r(\tau)]
\end{align}

The expectation of equation \ref{eq:likelihoodRatioExpectation} is useful to estimate the gradient of \(J_{\bm{\theta}}\) while avoiding the sum over all trajectories, which is intractable. The inner term \(\nabla_{\bm{\theta}}\log p_{\bm{\theta}}(\tau)r(\tau)\) still depends on the possibly unknown or not differentiable system dynamics, which now can be easily excluded using equation \ref{eq:probTrajectory}, because they do not depend on the parameters \(\bm{\theta}\):

\begin{align}
\nabla_{\bm{\theta}}\log p_{\bm{\theta}}(\tau)
&=\nabla_{\bm{\theta}}\log p(s_1)+\sum_{i=1}^{T-1}\nabla_{\bm{\theta}}\log \pi_{\bm{\theta}}(a_i|s_i)+\sum_{i=1}^{T-1}\nabla_{\bm{\theta}}\log p(s_{i+1}|s_i,a_i)\\
\label{eq:gradientEstimateTrajectory}
&=\sum_{i=1}^{T-1}\nabla_{\bm{\theta}}\log \pi_{\bm{\theta}}(a_i|s_i)
\end{align}

This means, all knowledge about the dynamics of the environment can be easily discarded to form a model-free estimate of the parameter gradient. In fact, the gradient can be approximated by sampling trajectories from the reinforcement learning environment to form a Monte-Carlo estimate yielding the REINFORCE learning rule (\cite{williams1992simple}). For any differentiable stochastic policy, it is straightforward to obtain an unbiased estimate of the gradient using this technique. Therefore, equation \ref{eq:rewardSignal} and \ref{eq:gradientEstimateTrajectory} are incorporated into equation \ref{eq:likelihoodRatioExpectation} with m being the number of sampled trajectories and \(T_i\) the length of i-th trajectory. \(s_j^i\) is the j-th state of the i-th trajectory, \(a_j^i\) the j-th action of the i-th trajectory and \(r_{j+1}^i\) the reward associated to both:

\begin{equation}
\label{eq:REINFORCE}
\nabla_{\bm{\theta}}J(\bm{\theta})\approx\frac{1}{m}\sum^{m}_{i=1}\sum^{T_i-1}_{j=1}\nabla_{\bm{\theta}}\log\pi_{\bm{\theta}}(a_j^i|s_j^i)\gamma^{j-1}r_{j+1}^i
\end{equation}

The original REINFORCE algorithm additionally uses a baseline term to reduce the variance of the gradient estimation. \textcite{williams1992simple} show, that the baseline term does not introduce a bias, if it is chosen independently from the selected actions. According to \textcite{degris2012model}, a reasonable choice for the baseline term is to use an estimate of the state value \(V_{\pi}(s_t)\). Equation \ref{eq:REINFORCE} with incorporated baseline becomes to:

\begin{equation}
\nabla_{\bm{\theta}}J(\bm{\theta})\approx\frac{1}{m}\sum^{m}_{i=1}\sum^{T_i-1}_{j=1}\nabla_{\bm{\theta}}\log\pi_{\bm{\theta}}(a_j^i|s_j^i)(\gamma^{j-1}r_{j+1}^i-b(s_j^i))
\end{equation}

\subsection{Actor-Critic Methods}
Section \ref{sec:likelihoodRatioMethods} has shown how to use the likelihood-ratio to estimate the gradient of the target function \(J\). \textcite{sutton2000policy} generalize this insight to the form of the \textit{policy gradient theorem}:

\begin{equation}
\label{eq:policyGradientTheorem}
\nabla_{\bm{\theta}}J(\bm{\theta})=\mathbb{E}_{\{s1,a1..s_T,a_T\}\sim p(\tau|\bm{\theta})}\Big[\sum_{i=1}^{T-1}\nabla_{\bm{\theta}}\log \pi_{\bm{\theta}}(a_i|s_i)Q^{\pi_{\bm{\theta}}}(s_i,a_i)\Big]
\end{equation}

Because there are many well studied methods to approximate value functions, for instance those described in section \ref{sec:valueFunctionAlgorithms}, it seems natural to use a second function estimator to also approximate a suitable value function to be used as training target for the policy gradient method. One advantage of this dual training approach in comparision to purely value based methods is, that the policy is still parametrized independently and thus can be used to output continuous actions. Other advantages of policy gradient based methods over value based methods, like making small updates to the parameters of the policy, also still apply when using estimates of a value function as training target.

\begin{figure}[ht]
\centering
\includegraphics[trim={0 0.8cm 0 0}, clip,width=\textwidth]{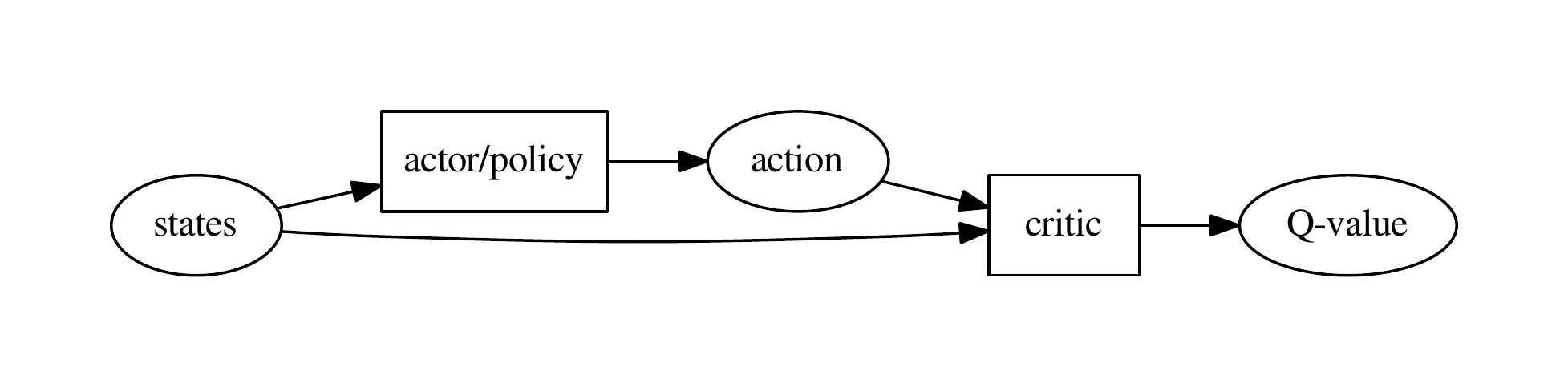}
\caption{Simple Actor-Critic architecture, where the critic estimates the action-value function. }
\label{fig:actorCritic}
\end{figure}

Methods that learn a value function, which is used as training target for an independently parametrized policy are called actor-critic architectures (\cite{sutton1998reinforcement}). The actor is the function estimator for the policy discussed so far. The critic is a second function estimator, that estimates a value function. Both parts can be modeled by neural networks. The interactions of the two function estimators are depicted in figure \ref{fig:actorCritic}. A straightforward way to derive an actor-critic like training method from the REINFORCE algorithm discussed in section \ref{sec:likelihoodRatioMethods} is to use the state-value function as baseline, which obviously requires a seperate estimate of a value function. \textcite{degris2012model} compare different actor-critic methods and show that they can be applied to solve a robotic task.

By approximizing the gradient of the true expected return by the gradient of a value function, a bias might be introduced. \textcite{sutton2000policy} show that under certain conditions, the gradient is still exact. This formulation is known as the \textit{policy gradient theorem with function approximation}.

In order to reduce the variance of the critic, it is very sensible not to approximize the Q-function, but another meaningful function, that can be used as target for policy optimization. For example, subtracting the state-value function \(V_{\pi_{\bm{\theta}}}(s_t)\) from \(Q_{\pi_{\bm{\theta}}}(s_t,a_t)\) leaves the policy gradient theorem intact, as \(V_{\pi_{\bm{\theta}}}(s_t)\) is independent from the selected action and thus does not alter the gradient with respect to the parameters of the policy. The term \(Q_{\pi_{\bm{\theta}}}(s_t,a_t)-V_{\pi_{\bm{\theta}}}(s_t)\), which is known as the advantage function \(A_{\pi_{\bm{\theta}}}(s_t,a_t)\) from section \ref{sec:dqnImprovements}, has very reduced variance however, as it only models the impact of the currently selected action on the expected return.

\subsection{Deep Deterministic Policy Gradient (DDPG)}
\label{sec:ddpg}

While it is possible to derive the gradient of a stochastic policy via the policy gradient theorem (equation \ref{eq:policyGradientTheorem}), it can be beneficial to use a deterministic policy for calculating the gradient.
\textcite{silver2014deterministic} show that the deterministic policy gradient is the expected gradient of the action-value function. They also prove, that the deterministic policy gradient is a special case of the gradient of many stochastic policies, when variance approaches zero. Let \(p(s|\mu_{\bm{\theta}})\) denote the probability of arriving in state s, when following the policy \(\mu_{\bm{\theta}}\). Because the policy is now a deterministic function, no trick is needed for differentiation and the policy gradient can be decomposed as follows:

\begin{align}
\nabla_{\bm{\theta}}J(\bm{\theta})&=\mathbb{E}_{s\sim p(s|\mu_{\bm{\theta}})}\big[\nabla_{\bm{\theta}}Q^{\mu_{\bm{\theta}}}(s,\mu_{\bm{\theta}}(s))\big]\\
&=\mathbb{E}_{s\sim p(s|\mu_{\bm{\theta}})}\big[\nabla_{\bm{\theta}}\mu_{\bm{\theta}}(s)\nabla_{\mu_{\bm{\theta}}(s)}Q^{\mu_{\bm{\theta}}}(s,\mu_{\bm{\theta}}(s))\big]
\label{eq:deterministicPolicyGradientTheorem}
\end{align}

As the expectation is taken only with respect to the states, it can be estimated more effectively than in the stochastic case, where the expectation depends on both the states and actions (see equation \ref{eq:policyGradientTheorem} for comparision). Obviously, the learning algorithm uses the gradient of the action-value function with respect to the action to improve the policy. Each training step modifies the policy in the way, that it's outputs are pushed in the direction of the positive gradient of the action-value function. Especially for continuous actions, this strategy is very effective, as it directly pushes the generated actions towards the assumed best action with respect to the action-value estimations. For a stochastic policy the same procedure would require a more exhaustive search in the action space to find the assumed best action.

\textcite{lillicrap2015continuous} apply these insights to problems with complex continuous action spaces and successfully combine the deterministic policy gradient with a deep Q-network (DQN) to obtain the deep deterministic policy gradient (DDPG) algorithm, that is shown in listing \ref{lst:ddpg}. To encourage exploration, a stochastic policy is still used to generate the training samples, which yields an off-policy training algorithm. Like DQN, the DDPG algorithm uses target networks for both the actor and the critic and experience replay. In contrast to DQN, the target networks are updated after each gradient step to slowly replicate the changes made to the trained networks.

\begin{minipage}{\linewidth}
\begin{lstlisting}[language=Matlab,mathescape,label={lst:ddpg}, caption={DDPG algorithm. Reproduced from \textcite{lillicrap2015continuous}. }, captionpos=t,backgroundcolor=\color{white},frame=lines]
Randomly initialize critic network $Q(s,a|\theta^Q)$ and actor $\mu(s|\theta^\mu)$ with weights $\theta^Q$ and $\theta^\mu$
Initialize target network $Q'$ and $\mu'$ with weights $\theta^{Q'}\leftarrow\theta^Q$, $\theta^{\mu'}\leftarrow\theta^\mu$
Initialize replay buffer $R$
for episode = 1, M do
    Initialize a random process $\mathcal{N}$ for action exploration
    Receive initial observation state $s_1$
    for t = 1, T do
        Select action $a_t=\mu(s_t|\theta^\mu)+\mathcal{N}_t$ according to the current policy and exploration noise
        Execute action $a_t$ and observe reward $r_t$ and observe new state $s_{t+1}$
        Store transition $(s_t,a_t,r_t,s_{t+1})$ in $R$
        Sample a random minibatch of $N$ transitions $(s_i,a_i,r_i,s_{i+1})$ from $R$
        Set $y_i = r_i + \gamma Q'(s_{i+1}, \mu'(s_{i+1}|\theta^{\mu'})|\theta^{Q'})$
        Update critic by minimizing the loss: $L=\frac{1}{N}\sum_i(y_i-Q(s_i,a_i|\theta^Q))^2$
        Update the actor policy using the sampled policy gradient:

            $\displaystyle\nabla_{\theta^\mu}J\approx\frac{1}{N}\sum_{i}\nabla_a Q(s,a|\theta^Q)|_{s=s_i,a=\mu(s_i)}\nabla_{\theta^\mu}\mu(s|\theta^\mu)|_{s=s_i}$

        Update the target networks:
        $\displaystyle\theta^{Q'}\leftarrow\tau\theta^Q+(1-\tau)\theta^{Q'}$
        $\displaystyle\theta^{\mu'}\leftarrow\tau\theta^\mu+(1-\tau)\theta^{\mu'}$
    end for
end for
\end{lstlisting}
\end{minipage}

\subsection{Asynchronous Advantage Actor-Critic (A3C)}
\label{sec:a3c}

The asynchronous advantage actor-critic algorithm (A3C), introduced by \textcite{mnih2016asynchronous} among other algorithms, is a popular recent implementation of an actor-critic model, that improves state-of-the-art performance on many experiments. The authors show for instance, that their algorithm is able to perform better than the deep Q-network, that is discussed in section \ref{sec:dqn}, on the task of playing many different Atari games. \textcite{levine2016learning} and \textcite{gu2017deep} show successful applications of asynchronous updates similar to those of A3C in robotics, where the algorithm is able to generalize to different robotic hardware. Another important benefit of the A3C algorithm is, that it can be efficiently implemented and is therefore faster than many other methods, that achieve comparable performance.

\begin{figure}[ht]
\centering
\includegraphics[trim={0 0.8cm 0 0}, clip, width=0.87\textwidth]{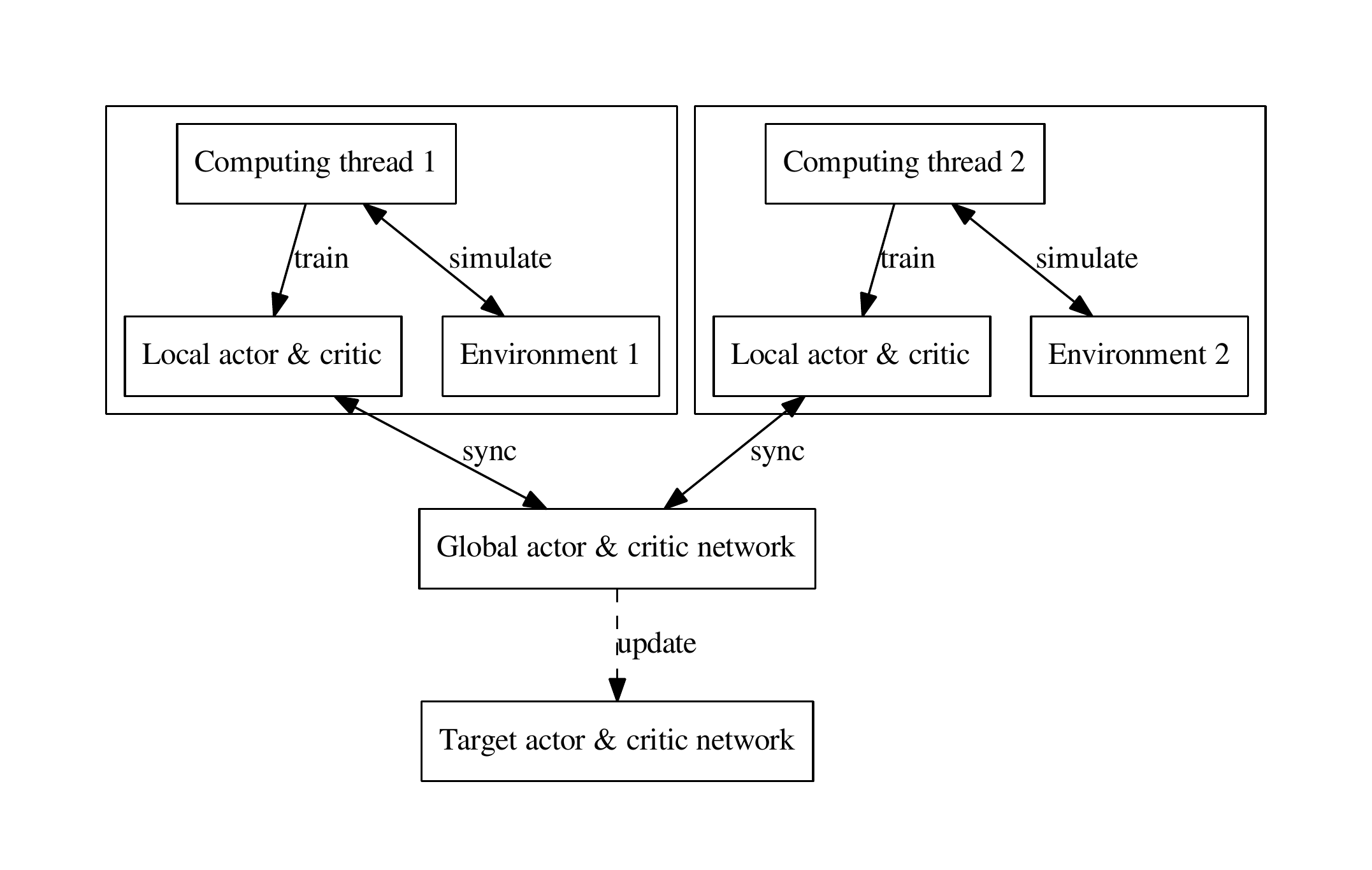}
\caption{Asynchronous advantage actor-critic (A3C) still uses target netorks for stability, but no experience replay. Many threads, each with a separate instance of the environment, train local instances of the actor and critic network in parallel, while only two threads are exemplary displayed. The local updates are synchronized with the global networks at regular intervals.}
\label{fig:a3c}
\end{figure}

The key idea that motivated asynchronous algorithms is that learning can be parallelized using different threads, that independently collect experience. The independent execution of multiple different environments reduces the variance of the trained estimators, because it provides the learning algorithm with many decorrelated training examples at one time. Techniques like experience replay used for training the DQN are thus no longer necessary. By choosing different starting conditions and exploration rates for the threads, it can be ensured that the training examples produced at one time are sufficiently varying. Figure \ref{fig:a3c} depicts the main components of the A3C algorithm.

Asynchronous advantage actor-critic and other similar algorithms like asynchronous n-step Q-learning (\cite{mnih2016asynchronous}) use a mix of explicitly computed n-step returns. Unlike algorithms that rely on eligibility traces, which were discussed in section \ref{sec:eligibilityTraces}, A3C directly executes a sequence of steps with fixed length n. After that, the one-step return is used to obtain the gradient update for the last pair of state and action, the two-step return for the second last and so on.

Each asynchronous thread independently computes updates for the parameters of both networks using the gradient of equation \ref{eq:policyGradientTheorem} to determine the update of the actor and the respective n-step update for updating the critic. Although there is a chance of overriding changes made by other threads, the gradient updates can be synchronized without any locks, if the learning rate is sufficiently small. This updating mechanism is known as Hogwild! style updating (\cite{recht2011hogwild}). Each thread maintains a local copy of the two global networks to being able to compute the updates independently from all other threads. After each update, the locally copied networks are updated. In comparision to other state-of-the-art methods, A3C is very fast, because of the possibility to massively parallelize it with minimal overhead for synchronization. Listing \ref{lst:a3c} pictures the algorithm in pseudocode.

\begin{minipage}{\linewidth}
\begin{lstlisting}[language=Matlab,mathescape,label={lst:a3c}, caption={A3C algorithm for each learner thread. The threads repeatedly synchronize their respective weight updates. Reproduced from \textcite{mnih2016asynchronous}. }, captionpos=t,backgroundcolor=\color{white},frame=lines]
// Assume global shared parameter vectors $\theta$ and $\theta_ v$ and global shared counter $T = 0$
// Assume thread-specific parameter vectors $\theta'$ and $\theta'_v$
Initialize thread step counter $t \leftarrow 1$
repeat
    Reset gradients: $d\theta\leftarrow 0$ and $d\theta_v\leftarrow 0$
    Synchronize thread-specific parameters $\theta'=\theta$ and $\theta'_v=\theta_v$
    $t_{start} = t$
    Get state $s_t$
    repeat
        Perform $a_t$ according to policy $\pi(a_t|s_t;\theta')$
        Receive reward $r_t$ and new state $s_{t+1}$
        $t\leftarrow t+1$
        $T\leftarrow T+1$
    until terminal $s_t$ or $t-t_{start} == t_{max}$
    $R = \begin{cases}
        0 &\text{for terminal}\,s_t\\
        V(s_t, \theta'_v) &\text{for non-terminal}\,s_t\,\text{// Bootstrap from last state}
    \end{cases}$
    for $i \in \{t-1, ... ,t_{start}\}$ do
        $R\leftarrow r_i+\gamma R$
        Accumulate gradients wrt $\theta':d\theta\leftarrow d\theta+\nabla\theta'\log\pi(a_i|s_i;\theta')(R-V(s_i;\theta'_v))$
        Accumulate gradients wrt $\theta'_v:d\theta_v \leftarrow d\theta_v+{\partial(R-V(s_i;\theta'_v))}/{\partial\theta'_v}$
    end for
    Perform asynchronous update of $\theta$ using $d\theta$ and of $\theta_v$ using $d\theta_v$
until $T > T_{max}$
\end{lstlisting}
\end{minipage}

\chapter{Extensions to Reinforcement Learning}
\label{ch:extensionsToRl}

Recent research often focuses on improving model-free reinforcement learning algorithms to being able to solve a wide variety of different challenging tasks with the same neural network structure and minimal or no changes to the used learning algorithms or hyperparameters (\cite{mnih2016asynchronous}). These methods, some of which were discussed at the end of section \ref{sec:rl}, have in common that they directly train a policy to solve the given task with one learning algorithm. The technique of learning a function approximator that directly predicts the desired output from the given input data without intermediate stages can be named \textit{end-to-end} learning (\cite{levine2016end}), mostly from pixels to actions. The execution of the policy during test time is straightforward and minimal or no preprocessing of the input data is required. While the obtained results are impressive, it is still reasonable to assume that incorporating information about the model of the environment or preprocessing the input data can improve the learned behavior.

As a practical example, \textcite{levine2016learning} use a training objective, which is similar to that of reinforcement learning, to predict the probability of successfully grasping an object, when the position of a robotic gripper is known. For testing the learned behavior, a sampling algorithm is used, that samples movements of the robot and evaluates them according to the predicted probability of a successful grasp. This approach incorporates knowledge of the environment into the training and testing process and thereby improves the performance of the robot after training and is able to generalize to different robotic hardware.

In the following, three possible ways to extend reinforcement learning algorithms will be discussed. These ideas require at least some knowledge about the environment or process the given information in several stages to produce the desired output and thus cannot be described as end-to-end learning algorithms. Nevertheless, we assume that some of the gained insights can be transfered to other similar tasks or environments with minor adaptions.

\section{Pretraining a State Model Using the Physical States}
\label{sec:pretrainingPhysicalStates}
Sometimes additional information can be observed during training for tasks that should be carried out while only observing pixel data in the test case. A robotic system for instance might have access to the physical states of the important components like positions of objects or parts of the robot during training. Reinforcement learning only on the physical states is usually much more effective than learning directly from pixels, because the possibly complicated task of extracting the necessary information from the high dimensional pixel data is no longer required.

In addition, it is also possible to learn a policy that is able to act on pixels while incorporating knowledge about the physical states to enhance the training performance. \textcite{levine2016end} use a dual training approach to train a neural network to directly predict robotic motor torques using pixel images as input and thus are able to form an end-to-end learning algorithm. The training procedure however does include the physical states of the robot and other objects to force the network to learn useful features.

\begin{figure}[ht]
\centering
\includegraphics[trim={0 0.8cm 0 0}, clip,width=0.95\textwidth]{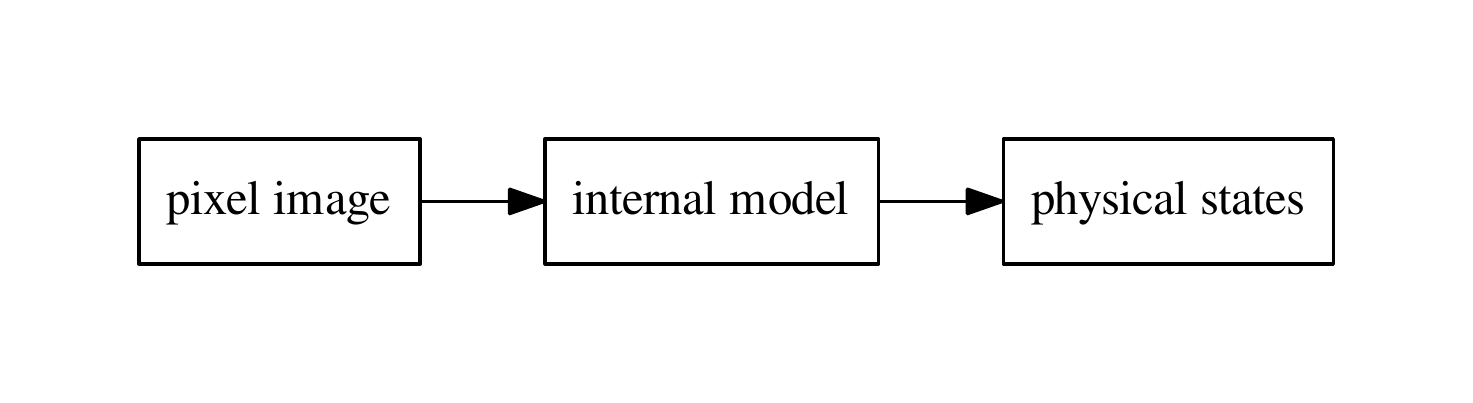}
\caption{The internal model predicts the physical states from a pixel image and thus enables any function estimator trained with reinforcement learning on the physical states to work on pixels.}
\label{fig:stateModelPhysicalStates}
\end{figure}

We suppose that introducing a model, that is trained to predict the physical states from a pixel image using a supervised learning algorithm (Figure \ref{fig:stateModelPhysicalStates}), shortens training time and simplifies the reinforcement learning task. We call this model an internal model. The term internal model sometimes refers to models learning the dynamics of the system (\cite{kawato1999internal}), while we label those inverse- or forward-models. The internal model can be trained in parallel to a second model, that uses reinforcement learning to predict actions based on the physical states. The resulting hybrid of the two parallel trained models cannot be described as an end-to-end model anymore, but it is still able to fulfill the same task like a model trained end-to-end from pixels to actions during test time as it predicts actions from pixels with an intermediate step in between but without any additional information required.

\section{Pretraining a State Model with a Deep Autoencoder}
\label{sec:pretrainingAutoencoder}

Without the knowledge of the physical states, it is more difficult to pretrain a model, that transforms the input pixels into an intermediate representation to be used as input for a reinforcement learning algorithm. Although no supervised learning target is available, unsupervised learning techniques can be applied to extract information inherent in the provided data.

\begin{figure}[ht]
\centering
\includegraphics[width=0.66\textwidth]{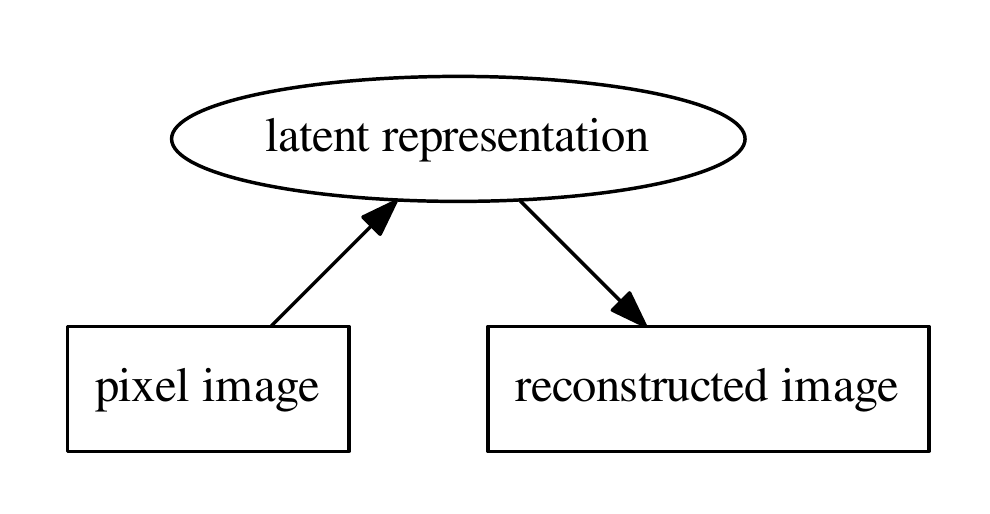}
\caption{Any autoencoder maps an input vector to a reconstruction through a latent internal representation. In the special case of using pixel images, the autoencoder is trained to reconstruct a given input image as precisely as possible.}
\label{fig:autoencoder}
\end{figure}

A convenient structure for learning to extract useful information from high dimensional data in an unsupervised manner is an \textbf{autoencoder} (Figure \ref{fig:autoencoder}), which is trained to reproduce its input through an internal representation or latent code (\cite{goodfellow2016deep}). In our case, the internal representation serves as input to the reinforcement learning process to form a hybrid training approach similar to the combination of internal model and reinforcement learning in section \ref{sec:pretrainingPhysicalStates}. The part of the autoencoder, that generates the latent code from the input is known as the \textit{encoder}, while the other part, that generates the reconstruction from the hidden code is named the \textit{decoder}. Both parts can be complicated function estimators, usually neural networks with multiple layers including convolutional layers and transposed convolutional layers.

To make the autoencoder learn something useful, it must be discouraged from simply copying the input to the output, which would make it useless. Building the latent code in fact must be tied to learning the important variations of the data. The autoencoder thus might not be able to recover the input exactly, but will focus on its main aspects. Autoencoders can be viewed as a feed forward network from the input to the reconstruction. Therefore, they can be trained by simple gradient descent similar to a supervised training procedure, where the input and the target are equal. The loss function to be optimized, that measures the difference of the input to the reconstruction in any way, is called the \textit{reconstruction error}.

\textcite{goodfellow2016deep} summarize several ways to train autoencoders. An autoencoder that is forced to learn useful information inherent in the data by making the dimension of its latent code smaller than the dimension of the input is called an \textbf{undercomplete autoencoder}. However, various regularization strategies enable autoencoders to still learn useful features of the data, when they are not necessarily undercomplete. This can be for example a sparsity contraint (\cite{lee2007efficient}) imposed on the latent code to form a \textbf{sparse autoencoder}. \textbf{Denoising autoencoders} add noise to the input and thus force the autoencoder to remove it and learn to distinguish realistic data from noise (\cite{vincent2008extracting}). Another strategy is to penalize the derivatives of the latent code with respect to the input to form a \textbf{contractive autoencoder}, which learns locally stable features (\cite{rifai2011contractive}).

Recent extensions to the general idea of learning from the task of reconstruction include deep autoencoders, that use deep neural networks to build the encoder and decoder functions. Another recent innovation in this field are autoencoders, that generalize the encoder and decoder functions to stochastic mappings. As popular examples of these probabilistic models, the adversarial autoencoder (\cite{makhzani2015adversarial}) and the variational autoencoder (\cite{kingma2013auto}) were successfully applied to a range of tasks including denoising, compression or semi-supervised classification, that is able to work on partially labled training data. Let \(p(\bm{x})\) denote the data distribution, \(\bm{\theta}\) the parameters of the encoder and \(\bm{\phi}\) the parameters of the decoder. The reconstruction error for both the encoder \(p_{encoder}(\bm{z}|\bm{x})=q_{\bm{\theta}}(\bm{z}|\bm{x})\) and the decoder \(p_{decoder}(\bm{x}|\bm{z})=p_{\bm{\phi}}(\bm{x}|\bm{z})\) being stochastic can be expressed as the negative log-likelihood of reconstructing x, when x is the input:

\begin{equation}
\mathcal{L}(\bm{x},\bm{\theta},\bm{\phi})_{rec}=\mathbb{E}_{\bm{z}\sim q_{\bm{\theta}}(\bm{z}|\bm{x})}\big[-\log(p_{\bm{\phi}}(\bm{x}|\bm{z}))\big]
\end{equation}

The \textbf{variational autoencoder} (\cite{kingma2013auto}) imposes a prior probability distribution on the latent code to regularize it and present the learned features in an appealing form that can be used to solve tasks like classification. The imposed prior usually is a multivariate gaussian disribution. The mean and variance of each variable in the latent code are modeled seperately and forced to match the prior distribution by adding a second term to the loss function beside the reconstruction error. The newly introduced term measures the difference between the actual probability distribution of the latent variables and the imposed prior \(p(\bm{z})\) using the Kullback–Leibler divergence:

\begin{equation}
\mathcal{L}(\bm{x},\bm{\theta},\bm{\phi})=\mathcal{L}(\bm{x},\bm{\theta},\bm{\phi})_{rec}+D_{KL}[q_{\bm{\theta}}(\bm{z}|\bm{x})||p(\bm{z})]
\end{equation}

\begin{figure}[ht]
\centering
\includegraphics[trim={0 1cm 0 0}, clip,width=0.68\textwidth,]{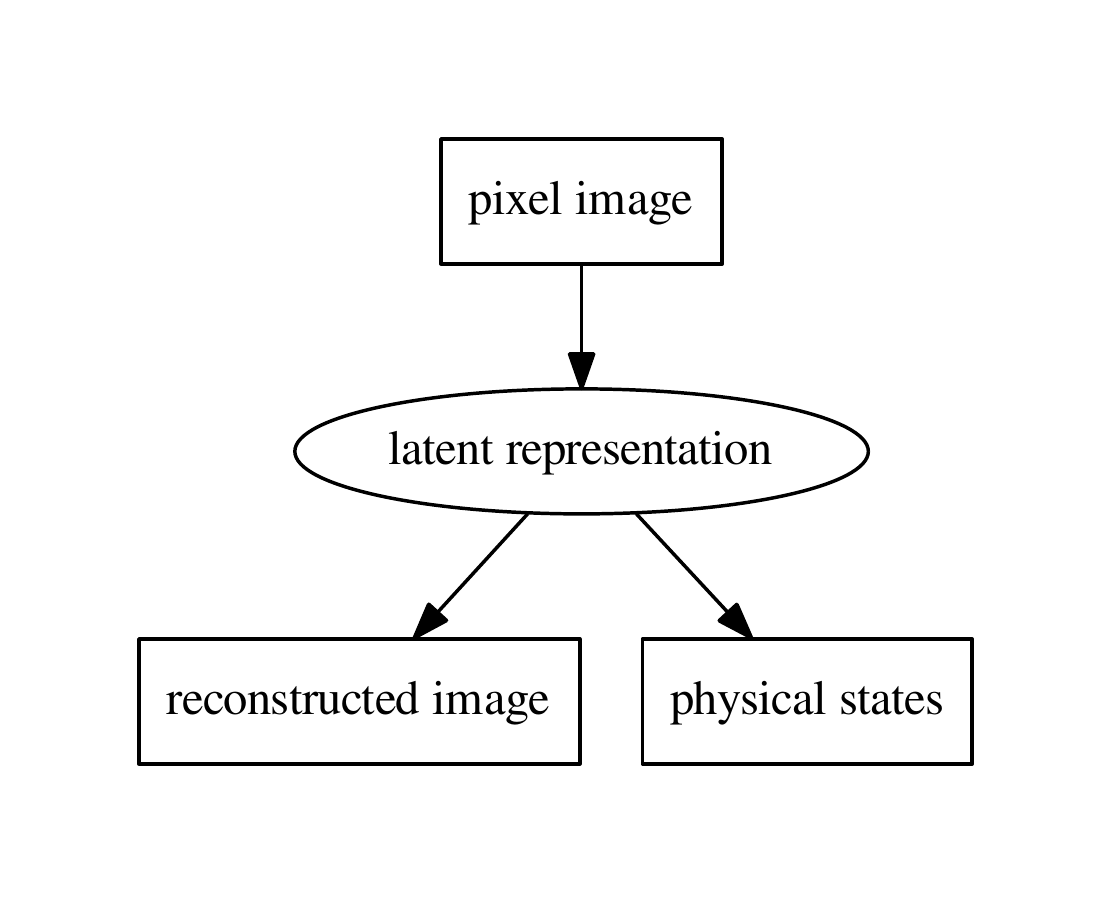}
\caption{A network structure similar to that of an autoencoder can be trained jointly to predict the reconstructed image and the physical states. The learned latent representation ideally encodes the physical states with less error than the purely supervised model of section \ref{sec:pretrainingPhysicalStates}. }
\label{fig:regularizedAutoencoder}
\end{figure}

In addition to a training a deep autoencoder only on pixel data, in the case when physical states are available, they might be incorporated into the training procedure of the autoencoder to help extracting sensible features. The structure of the autoencoder can be extended to not only predict the reconstructed image from the latent variables, but also the physical states (Figure \ref{fig:regularizedAutoencoder}). The training objective is to minimize the sum of the two losses for the reconstructed image and the predicted physical states respectively. To enhance the regularization effect, the latent variables of the autoencoder can be forced to become a linear function of the physical states, by drawing only simple linear connections between the two layers. When the dimension of the latent code is chosen to be larger than the dimension of the physical states, this learning technique might lead to a better estimation of the true physical states, because reducing the high dimensional pixel image to very few physical states can be error-prone and inexact.

\section{Training Inverse- or Forward-Models}
\label{sec:inverseForwardModels}

Reinforcement learning algorithms often do not require to directly learn a model of the system dynamics. Nevertheless they must maintain an implicit understanding of those dynamics to being able to act. Especially for end-to-end training approaches it is not easy to tell what kind of knowledge the trained model has aquired about its environment. Most of the time, only the actions taken and thus the behavior can be evaluated. Other recent innovations that are only loosely coupled to reinforcement learning but solve very similar problems directly attempt to learn the dynamics of the environment.

\begin{figure}[ht]
\centering
\includegraphics[trim={0 1cm 0 0}, clip,width=0.8\textwidth,]{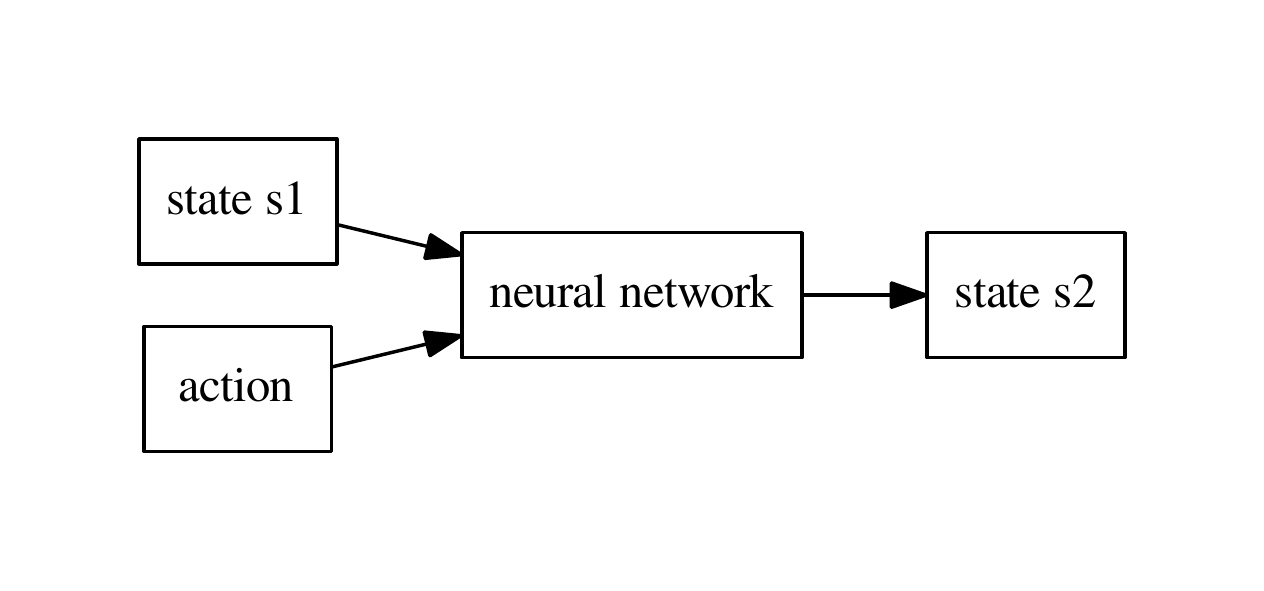}
\caption{Abstract architecture of a simple forward model.}
\label{fig:simpleForwardModel}
\end{figure}

\begin{figure}[ht]
\centering
\includegraphics[trim={0 1cm 0 0}, clip,width=0.8\textwidth,]{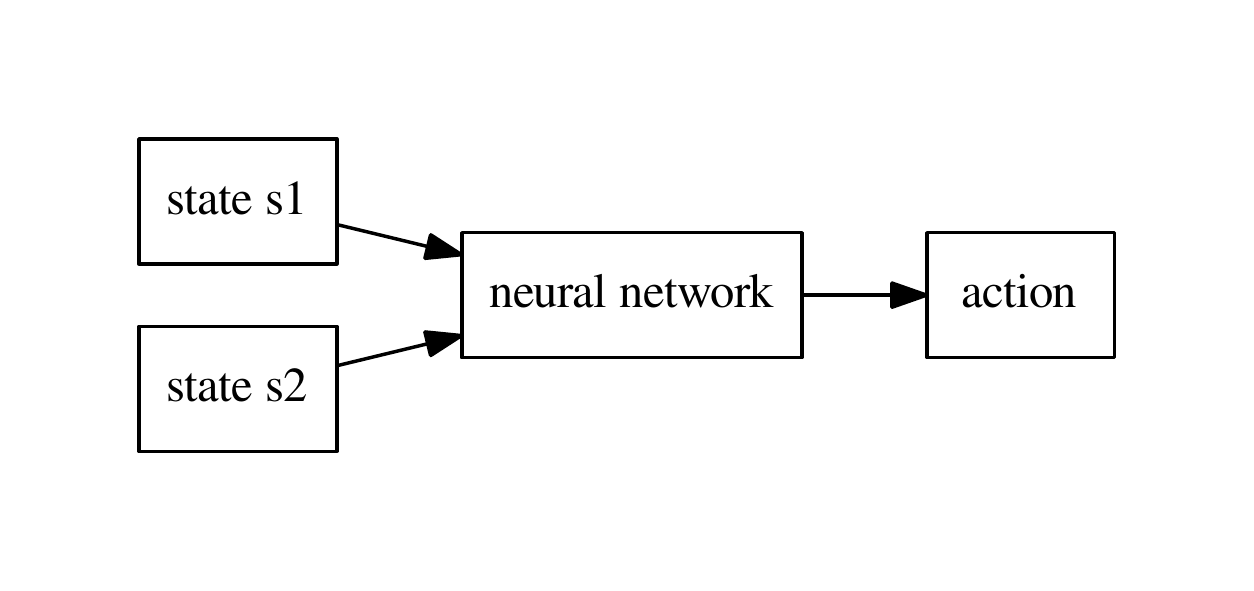}
\caption{Abstract architecture of a simple inverse model.}
\label{fig:simpleInverseModel}
\end{figure}

There are two main possibilities to learn the system dynamics \(p(s_{t+1}|s_t,a_t)\). While looking into the future, \textbf{forward models} (Figure \ref{fig:simpleForwardModel}) can be trained to predict the next state \(s_{t+1}\). Often it is also helpful to estimate the action that was executed between two states. Models of this kind are called \textbf{inverse models} (Figure \ref{fig:simpleInverseModel}). \textcite{kawato1999internal} introduce both types of models and discuss possible applications to motor control problems.

Both approaches usually rely on neural networks as function approximators, when they are used in settings similar to the reinforcement learning tasks described so far. \textcite{dosovitskiy2016learning} train a forward model to being able to compare a set of discrete actions by evaluating their respective impact in the context of the 3D game Doom. \textcite{agrawal2016learning} jointly train a forward and an inverse model in a robotic environment to intuitively learn about the dynamics of different physical objects by displacing or rotating them.

\chapter{Methods}
\label{ch:methods}

An important property of many recent inventions in the field of reinforcement learning is, that the underlying ideas can be transferred to many different environments with minimal changes. In contrast to this, to evaluate the different reinforcement learning methods discussed so far and extensions to them, we focus on a single simulated robotic task for two reasons. First, we assume that the relative scores obtained by training a single robotic task with different algorithms are still meaningful and can be used to compare the algorithms, while training on a larger set of tasks would require a more sophisticated design of the
function estimators beforehand and thus would slow down the experiments. Second, solving a single task allows to incorporate information about the model of the environment into the training algorithm, as discussed in section \ref{ch:extensionsToRl}. An important point to investigate is the influence of this additional knowledge on the training performance. Recently published ideas, that analyse the benefits of directly modeling the underlying system dynamics, for instance the algorithm proposed by \textcite{dosovitskiy2016learning}, are often tested on a very small set of similar environments too.

In the following, we will first describe the single simulated robotic task used for all experiments and its variations. Then follows a discussion of the software components used for conducting the experiments. Finally we summarize the different algorithms that were used. Two new algorithms, that we call distributed and asynchronous DDPG will be presented. With these training approaches, we aim to combine asynchronous methods and DDPG to form asynchronous learning methods, which are able to effectively learn to predict continuous actions.

\section{Training Environments}
\label{sec:trainingEnvironments}
We trained all experiments on a single robotic task, that is very similar to the OpenAI Gym reacher task\footnote{\url{https://gym.openai.com/envs/Reacher-v1}, last downloaded 2017-09-14}. OpenAI Gym\footnote{\url{https://gym.openai.com/}, last downloaded 2017-09-14} is an open source platform for comparing reinforcement learning algorithms. It is used to obtain benchmark results for algorithms trained on many different environments like video games, robotic tasks or board games through a unified programming interface.

The reacher task consists of a simulated arm with two degrees of freedom, where the gripper of the robotic arm, which is the endpoint of the last arm segment, should be guided to reach a target. While it would be possible to transfer the reacher problem to a 3D space with more degrees of freedom, we restrict to the 2D space for all experiments and only use two arm segments.

The original reacher task provides the learning algorithm with the physical states, which resemble the angles of the arm segments and the position of the target, during training and test time. In contrast to this, we try to learn a policy, that is able to predict actions from pixel data. We therefore provide most experimentally learned policies only with screenshots of the simulated reacher task during test time. This task is much harder, as the state space is very much larger than before and the tested algorithm needs to find a way to extract useful information from pixels. Ideally, we would like to be able to train this kind of policy only on pixel data and thus leave out the physical states completely. Nevertheless, many experiments still include the information about the physical states as guide for the algorithm to help it extracting useful information from the pixels.

\subsection{Simulation with Matplotlib}
\begin{figure}[ht]
\centering
\includegraphics[width=0.8\textwidth]{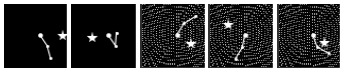}
\caption{Five visualizations of the matplotlib reacher at different random states. The first two images are rendered without background noise, while the other images include static noise in the background.}
\label{fig:matplotlibEnvironment}
\end{figure}

As the OpenAI Gym reacher environment relies on the commercial mujoco physics engine, we decided to replicate the simulation in essence using matplotlib\footnote{\url{https://matplotlib.org/}, last downloaded 2017-09-14}. Figure \ref{fig:matplotlibEnvironment} shows five images, that were randomly generated using the matplotlib simulation. To make the vision task harder, which means making it harder to extract useful information from pixels, noise can be added in the background. The noise is however static and thus it looks always the same on every pixel image.

\begin{figure}[ht]
\centering
\includegraphics[width=0.35\textwidth]{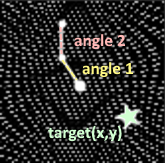}
\caption{Visualization of the reacher environment with highlighted and labeled physical states. The simulation relies on matplotlib. The arm segments are simple lines with dots in between. The target has the shape of a star. Noise in the background can be added in form of white dots.}
\label{fig:matplotlibPhysicalStates}
\end{figure}

The rendered images have 64 pixels in both dimensions. The environment outputs the physical states in parallel, which consist of the two angles of the arm segments and the coordinates of the goal. Figure \ref{fig:matplotlibPhysicalStates} shows a screenshot of the environment with the physical states highlighted. Executed actions are tuples with two components specifying the desired rotation of both arm segments. Each arm can rotate maximally by 2 degrees in each direction. The components of the actions are clipped to lie in the interval between -1 and 1: \(a\in[-1,1]^2\). Action (-1,-1) for instance means to rotate both segments left by two degrees and action (1,-1) means rotating only the first segment right by two degrees and the second segment left by two degrees.

The reward returned by the environment for each step is designed to guide the gripper to the target as fast as possible. We therefore introduced a distance related part of the reward function, that depends on the distance between the gripper and the target: \(r_{dist}=e^{-|x_{gripper}-x_{target}|}\). Early experiments showed that this kind of reward leads to a strategy, where the gripper circles around the target without ever reaching it, while collecting a large amount of reward, because the distance is very small. To overcome this problem, we multiplied the distance based reward with a control term, that only depends on the currently executed action: \(r_{ctrl}=|x_{gripper-old}-x_{target}|-|x_{gripper}-x_{target}|\), which forces the action to be taken in the direction of the target. As the total reward \(r=r_{dist}\cdot r_{ctrl}\) is the product of both terms, actions that are taken in the wrong direction, when being close to the target, will be severely punished. The total reward is normalized to lie in the interval [-1,1].

When \(r_{dist}\) falls below 0.1 the episode is considered successfully terminated. This distance is sufficiently small for reaching the target, as both axes of the simulated environment range from -1 to 1. The episode fails, when 1000 steps are executed without reaching the target.

\subsection{More Realistic Simulation with Dart}
\label{sec:dartEnvironment}

\begin{figure}[ht]
\centering
\includegraphics[width=0.5\textwidth]{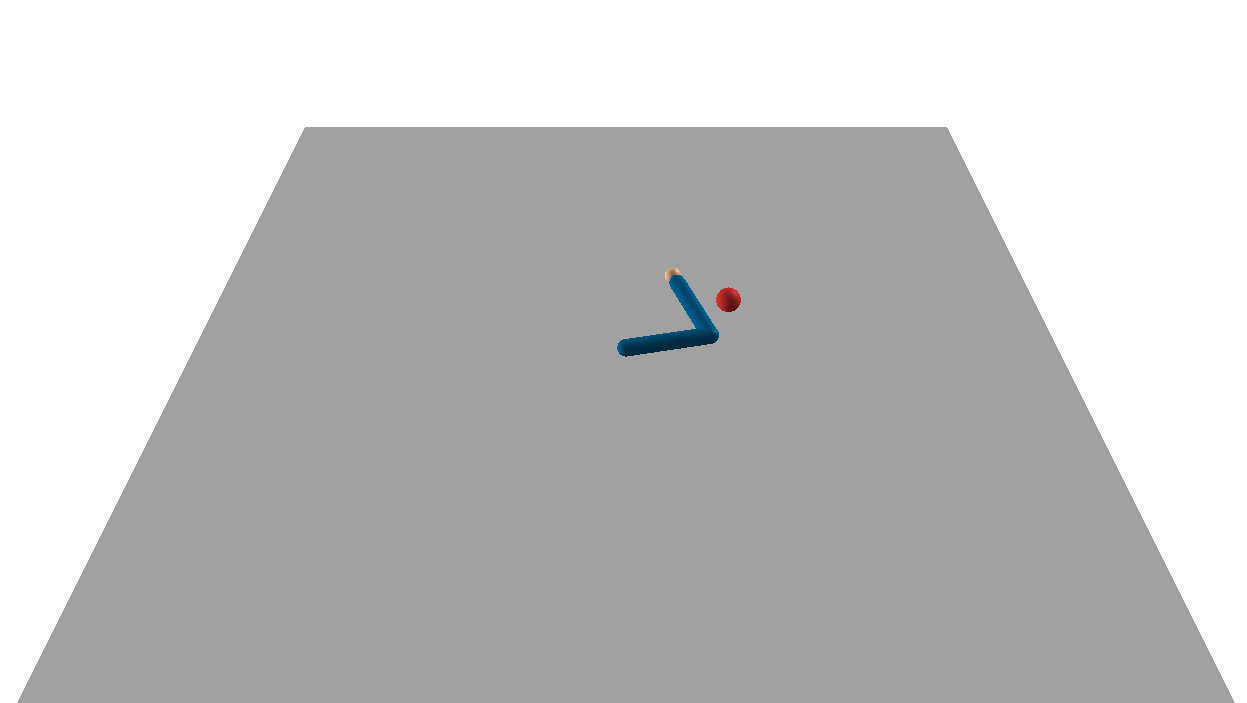}
\caption{Screenshot of the original visualization of the Dart environment, which simulates the reacher task in 3D, where the arm is only allowed to move on a two-dimensional plane. The target point is already enlarged here.}
\label{fig:dartReacher}
\end{figure}

The same task of reaching a target with a robotic arm, that has two degrees of freedom, has been replicated as an extension to OpenAI Gym using the open source physics engine Dart\footnote{\url{https://github.com/DartEnv/dart-env}, last downloaded 2017-09-14}. We also experimented with this visualization to obtain more realistic pixel images. The simulation with Dart has however not been designed to allow training from pixel data. Hence, we had to apply preprocessing to the pixel images, which means that we enlarged the target point to make it better recognizable by the training algorithm. Then we cropped the important region of the image, where the arm and the target are shown and resized the cropped images to 64x64 pixels. The physical states were adjusted to be similar to those of the matplotlib simulation for better comparision. Instead of the 11 physical states originally returned by the Dart environment, we used these to calculate four variables representing the angles and the target position, that are finally passed to the learning algorithm.

\begin{figure}[ht]
\centering
\includegraphics[width=0.8\textwidth]{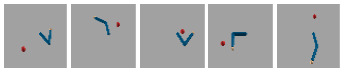}
\caption{Five visualizations of the Dart reacher at different random states. The images have been cropped to the important region and resized to 64x64 pixels.}
\label{fig:dartResized}
\end{figure}

\section{Implementation Details}
\subsection{Used Software and Hardware}
All implementations purely consist of python code, while numpy is used for general data processing. We used tensorflow\footnote{\url{https://www.tensorflow.org/}, last downloaded 2017-09-14} for neural network training and keras\footnote{\url{https://keras.io/}, last downloaded 2017-09-14} to build the architecture of the respective networks. Keras can be seen as an easy interface to tensorflow, while it also supports other backends. For drawing plots and statistics we relied on matplotlib\footnote{\url{https://matplotlib.org/}, last downloaded 2017-09-14} and tensorboard, which is a part of the tensorflow framework and provides easy to use visualization tools for tensorflow training and neural network training architectures (\textit{graphs}).

With tensorflow it is relatively easy to set up an asynchronous training algorithm, as the framework itself heavily supports multithreading and is even able to run on a GPU without any changes to the code of the algorithm. When executed without any restrictions, Tensorflow occupies all computing resources it can find, which means it creates a computing thread for every CPU and also reserves the whole memory of all available GPUs. For CPU training, tensorflow distinguishes between intra-op-parallelism and inter-op-parallelism. For a single computing operation, the threads of the intra-op-parallelism pool are used to execute this operation in parallel, while the threads in the inter-op-parallelism pool execute multiple operations at one time. For asynchronous training, especially the inter-op-parallelism is important, as multiple threads independently compute gradients, which corresponds to executing multiple gradient operations in parallel. The \texttt{intra\_op\_parallelism\_threads} variable and the \texttt{inter\_op\_parallelism\_threads} variable control the size of the respective thread pools. Listing \ref{lst:tensorflowParallelism} shows a sample configuration of both variables.

\begin{minipage}{\linewidth}
\begin{lstlisting}[language=Python,label={lst:tensorflowParallelism}, caption={An example of how to specify the size of the tensorflow thread pools at the beginning of the python program. Both sizes are exemplary set to 4. The newly created tensorflow session should be set as default session for the backend of the keras library.}, captionpos=t,backgroundcolor=\color{white},frame=lines,keywordstyle=\bfseries]
import tensorflow as tf
from keras import backend as K

def main(_):
    intra_threads = 4
    inter_threads = 4
    sess = tf.Session(config=tf.ConfigProto(intra_op_parallelism_threads=intra_threads, inter_op_parallelism_threads=inter_threads))
    with sess.as_default():
        K.set_session(sess)
        # do calculations ...
\end{lstlisting}
\end{minipage}

To restrict tensorflow to use only one GPU of a cluster, the \texttt{CUDA\_VISIBLE\_DEVICES} switch is necessary to specify the indices of the visible GPUs. For most cases it makes sense to train on a single GPU, while multiple indices could be set by using a colon as delimiter. The switch can be set for each run separately, e.g. \texttt{CUDA\_VISIBLE\_DEVICES=0 python main.py} for a file named \texttt{main.py}. Listing \ref{lst:gpuParameter} depicts how to rewrite the main function to support passing the GPU index as command line argument for easier use. The equivalent to directly using the switch would then be the shorter form: \texttt{main.py -\,-gpu=0}.

\begin{minipage}{\linewidth}
\begin{lstlisting}[language=Python,label={lst:gpuParameter}, caption={Initializing the \texttt{CUDA\_VISIBLE\_DEVICES} switch in code from a command line parameter is a practical solution to prevent tensorflow from unnecessarily blocking the memory of all GPUs. If an invalid index (e.g. -1) is specified, tensorflow runs only on the CPUs.}, captionpos=t,backgroundcolor=\color{white},frame=lines,keywordstyle=\bfseries]
import tensorflow as tf

flags = tf.app.flags
flags.DEFINE_integer('gpu', 0, 'index of GPU to use')
FLAGS = flags.FLAGS

def main(_):
    os.environ["CUDA_VISIBLE_DEVICES"] = str(FLAGS.gpu)
\end{lstlisting}
\end{minipage}

It is also possible to assign a specific device to a subset of tensorflow operations. The block, where the operations are defined has to start with a wrapping call to \texttt{tf.device(device)}. Listing \ref{lst:tensorflowDevice} shows this method for a simple example. We were however unable to achieve any performance improvements when directly placing operations on devices and thus decided to let tensorflow automatically assign the operations to the available hardware for all experiments.

\begin{minipage}{\linewidth}
\begin{lstlisting}[language=Python,label={lst:tensorflowDevice}, caption={Tensorflow operations can be directly assigned to hardware units, for example the CPU with index 1.}, captionpos=t,backgroundcolor=\color{white},frame=lines,keywordstyle=\bfseries]
import tensorflow as tf

with tf.device('/cpu:1'):
    # all operations defined here will run on CPU 1

\end{lstlisting}
\end{minipage}

For training on CPUs we used a computing server with 4x8 Intel Xeon E5-4650 CPUs and 256 GB RAM. The GPU based training was performed on a second machine with a shared-memory system, 16 CPU cores, 8 Tesla K20m GPUs and 126 GB RAM. The reinforcement learning algorithms were run only on the CPUs without exception. As they require frequent interactions with the environment, which requires communication with the CPU, there were no speed gains observed when using a GPU. Especially asynchronous implementations can be efficiently run on a parallel CPU system, because execution can be carried out by multiple parallel threads, that only require minimal communication. All different forms of pretraining, which do not involve reinforcement learning, were entirely performed on the GPUs and heavily benefited from the speedup. Especially implementations that use large amounts of training data at once can be efficiently parallelized on a GPU.

\subsection{Asynchronously Executing Multiple Environments}
\begin{figure}[ht]
\centering
\includegraphics[width=\textwidth]{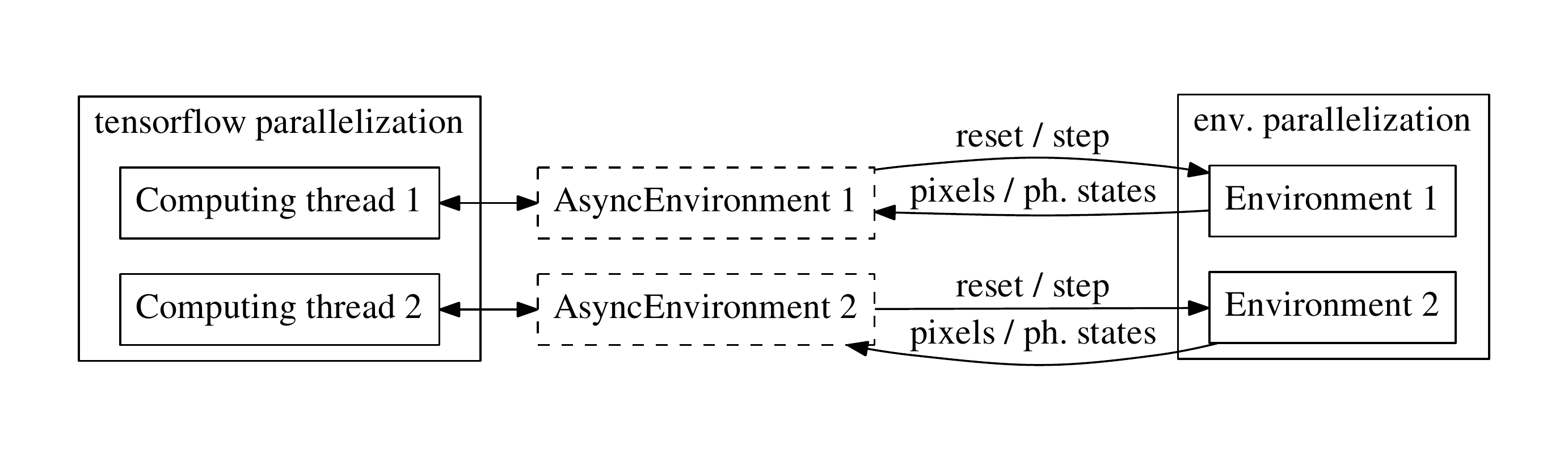}
\caption{The execution of different instances of the environment cannot be automatically parallelized by tensorflow. We therefore create a separate process for each instance of the environment and attach it to the computing thread.}
\label{fig:asyncEnvironments}
\end{figure}

For asynchronous training, we identified the execution of the environment being a major bottleneck, when using a single threaded instance of the environment for asynchronous training algorithms. Tensorflow takes care of parallelizing the network training operations, but does not parallelize the executions of the environment, which run synchronously due to pythons global interpreter lock\footnote{\url{https://wiki.python.org/moin/GlobalInterpreterLock}, last downloaded 2017-09-14}. We thus decided to use the \texttt{multiprocessing} module to create separate processes for all instances of the environment. Listing \ref{lst:asyncEnvironment} shows the snippet where a new simulation process is initialized. Figure \ref{fig:asyncEnvironments} shows how the \texttt{AsyncEnvironment} class is used, that provides an interface for creating and managing asynchronous environments. For easy usage, this class exposes the same function like a local environment, but sends the commands across process boundaries instead of directly executing them.

It is important to notice, that the creation mode of the new process must be set to \texttt{spawn}. The standard setting is \texttt{fork} on Unix/Linux platforms, which copies the entire memory of the computing process to all simulating processes and leads to explosion of memory. This happens especially, when the parameters of the neural networks, which are stored for the computing process, already consume large amounts of memory.

Another problem was that we were unable to simulate the Dart environment discussed in section \ref{sec:dartEnvironment} on a computing server, because the simulation is carried out with OpenGL and needs an active display to function properly. To separate the simulation from the training algorithm, we decided to set up a TCP server for simulation.

Especially for running asynchronous algorithms with multiple independent instances of the environment, it is crucial for the server to be able to effectively manage multiple client connections in parallel. Each client requests a unique id, which is bound to a single instance of the environment on the server side. The client then sends the actions to be executed and a reset command at the start of each episode to the server, while receiving the current pixel image and the physical states in return.

The architecture of the simulation server is thus very similar to the asynchronously simulated environments depicted in figure \ref{fig:asyncEnvironments}. The only difference is, that the communication with the environments is now TCP-based and carried out over the network.

\begin{minipage}{\linewidth}
\begin{lstlisting}[language=Python,label={lst:asyncEnvironment}, caption={The class AsynchronousEnvironment encapsulates the creation of separate processes with \texttt{multiprocessing}. The \_\,\_init\_\,\_ method starts a new process without copying the memory of the current process to it. The processes use pipes to communicate.}, captionpos=t,backgroundcolor=\color{white},frame=lines,keywordstyle=\bfseries]
import multiprocessing

class AsynchronousEnvironment:
    def __init__(self, env_name):
        ctx = multiprocessing.get_context('spawn')
        sim_pipe, self.pipe = ctx.Pipe()
        self.proc = ctx.Process(
            name="data_generator",
            target=func_proc,
            args=(sim_pipe,env_name))
        self.proc.start()
\end{lstlisting}
\end{minipage}

\section{Training Algorithms}
\label{sec:trainingAlgorithms}
As the robotic task we were aiming to solve uses continuous actions, we mainly focused on the DDPG algorithm (\cite{lillicrap2015continuous}), that was introduced in section \ref{sec:ddpg}, and deterministic policies. We use the DDPG implementation from keras-rl\footnote{\url{https://github.com/matthiasplappert/keras-rl}, last downloaded 2017-09-14} as a basline for our own experiments, but also provide an own implementation of two extended DDPG algorithms, that combine the ideas of asynchronous training algorithms like A3C (\cite{mnih2016asynchronous}; see section \ref{sec:a3c}) with the deterministic policy gradient. As this combination is a novel approach, one aim of the conducted experiments was to evaluate the performance of the extended DDPG algorithms.

We first investigated a variant of DDPG with one-step updates, that directly bootstrap from the next state (see section \ref{sec:eligibilityTraces}). This algorithm does not use Hogwild! style updates (\cite{recht2011hogwild}) to synchronize the updates of the different threads, but locks the weights of the networks during training for other threads. The algorithm executes five steps before performing a gradient update and stores the experience collected so far in a very small local memory. We therefore suggest to view this method as implementing a distributed experience replay memory rather than a fully asynchronous algorithm and call it \textbf{distributed DDPG}. This idea can be seen as an intermediate step between using experience replay and fully asynchronous updates. We still used target networks and updated them in the same way like proposed for plain DDPG (\cite{lillicrap2015continuous}).

Subsequent experiments also included a variant of DDPG with lock-free Hogwild! style updates, that we call \textbf{asynchronous DDPG}. We also switched to using the same mix of explicitly computed n-step returns like A3C (see section \ref{sec:a3c}).

\begin{minipage}{\linewidth}
\begin{lstlisting} [captionpos=t,backgroundcolor=\color{white},frame=lines, caption={Distributed DDPG algorithm for each actor thread with globally shared counter T and globally shared parameter vectors \(\theta_{\mu}\), \(\theta_{Q}\), \(\theta_{\mu}^{-}\) and \(\theta_{Q}^{-}\).}, label={lst:asyncDDPG}, language=Matlab,mathescape]
while $T < T_{max}$ do
    t = 0
    Get state $s_t$
    repeat
        Execute action $\displaystyle a_t$ according to policy $\displaystyle\mu(s_{t};\theta_{\mu})+\epsilon\mathcal{N}$
        Receive reward $\displaystyle r_{t+1}$ and observe new state $\displaystyle s_{t+1}$
        Compute $\displaystyle R_t = \begin{cases}
                r_{t+1} + \gamma Q(s_{t+1},\mu(s_{t+1};\theta_{\mu}^{-});\theta_Q^{-}) \, \text{if not terminal} \\
                r_{t+1} \, \text{otherwise.} \\ \end{cases}$
        Store $\displaystyle(s_t, a_t, R_t)$ in buffer
        if $t \,\%\, 5 == 0$ or terminal then
            Update critic using gradient: $\displaystyle \sum_{i=1}^5 \nabla_{\theta_{Q}} (R_i - Q(s_i, a_i;\theta_{Q}))^2$
            Update actor using gradient: $\displaystyle \sum_{i=1}^5 \nabla_{\mu(s_{i};\theta_{\mu})} Q(s_i, \mu(s_{i};\theta_{\mu});\theta_{Q}) \nabla_{\theta_\mu} \mu(s_i;\theta_{\mu})$
            Update target critic: $\displaystyle\theta_Q^{-}\leftarrow\tau\theta_{Q}+(1-\tau)\theta_Q^{-}$
            Update target actor:  $\displaystyle\theta_{\mu}^{-}\leftarrow\tau\theta_{\mu}+(1-\tau)\theta_{\mu}^{-}$
            Empty buffer
        end if
        $t = t + 1$
        $T = T + 1$
    until $t > 1000$ or terminal
\end{lstlisting}

\begin{lstlisting} [captionpos=t,backgroundcolor=\color{white},frame=lines, caption={Asynchronous DDPG algorithm for each actor thread with globally shared parameter vectors \(\theta_{\mu}\), \(\theta_{Q}\), \(\theta_{Q}^{-}\), \(\theta_{\mu}^{-}\) and counter T. \(\theta'_{\mu}\) and \(\theta'_{Q}\) are thread-specific copies. }, label={lst:asyncDDPGHogwild}, language=Matlab,mathescape]
while $T < T_{max}$ do
    t = 0
    Get state $s_t$
    repeat
        Execute action $a_t$ according to policy $\displaystyle\mu(s_{t};\theta'_{\mu})+\epsilon\mathcal{N}$
        Receive reward $r_{t+1}$ and observe new state $s_{t+1}$
        Store $\displaystyle(s_t, a_t, r_{t+1})$ in buffer
        $t = t + 1$
        $T = T + 1$
        if $t \,\%\, 5 == 0$ or terminal then
            $\displaystyle R = \begin{cases}
                0 &\text{for terminal}\,s_t\\
                Q(s_{t+1},\mu(s_{t+1};\theta_{\mu}^{-});\theta_Q^{-}) &\text{for non-terminal}\,s_t\
            \end{cases}$
            for $i \in \{t-1, ... ,t-5\}$ do
                $R\leftarrow r_i+\gamma R$
                Accumulate gradients wrt $\theta'_{Q}$ and $\theta'_{\mu}:$
                    $\displaystyle d\theta_{Q}\leftarrow\nabla_{\theta'_{Q}} (R - Q(s_i, a_i;\theta'_{Q}))^2$
                    $\displaystyle d\theta_{\mu}\leftarrow\nabla_{\mu(s_{i};\theta'_{\mu})} Q(s_i, \mu(s_{i};\theta'_{\mu});\theta'_{Q}) \nabla_{\theta'_\mu} \mu(s_i;\theta'_\mu)$
            end for
            Perform asynchronous update of $\theta_Q$ using $d\theta_Q$ and of $\theta_{\mu}$ using $d\theta_{\mu}$
            Update target critic: $\displaystyle\theta_{Q}^{-}\leftarrow\tau\theta_{Q}+(1-\tau)\theta_{Q}^{-}$
            Update target actor:  $\displaystyle\theta_{\mu}^{-}\leftarrow\tau\theta_{\mu}+(1-\tau)\theta_{\mu}^{-}$
            Reset gradients: $d\theta_{Q}\leftarrow 0$ and $d\theta_{\mu}\leftarrow 0$
            Synchronize thread-specific parameters $\theta'_{Q}\leftarrow\theta_{Q}$ and $\theta'_{\mu}\leftarrow\theta_{\mu}$
            Empty buffer
        end if
    until $t > 1000$ or terminal
\end{lstlisting}
\end{minipage}

\chapter{Experimental Results}
\label{ch:experimentalResults}

The experimental analysis aims to evaluate two different ideas. First, we compare the distributed and asynchronous DDPG algorithms to a standard DDPG baseline. Second, we investigate the effect of various ways to incorporate knowlege of the environment into the training process. The final goal is to build a training algorithm, that works only on pixel data during training and test time, although this algorithm does not necessarily have to be an end-to-end reinforcement learning algorithm. In the last section of the experiental analysis we will focus on ideas to apply preprocessing to pixel images without any knowledge of the physical states. The sections before will include the physical states either only during the training process or for both training and testing.

Due to computational limitations, it was not possible to perform an extensive hyperparameter search for any of the provided experiments. We therefore reused many hyperparameters and some elements of the network structure from \textcite{lillicrap2015continuous} and \textcite{mnih2016asynchronous}. For comparing the different variants of the DDPG algorithm, the same hyperparameters and network structure were used to obtain a reliable relative performance. We used the following hyperparameters for all experiments:

\begin {table}[ht]
\centering
\begin{tabular}{ l | l }
    parameter & value \\
    \hline
    update rate for target networks \(\tau\) & 0.001 \\
    learning rate critic & 0.0001 \\
    learning rate actor & 0.0001 \\
    discount factor \(\gamma\) & 0.97 \\
    weight penalty l2 (for critic weights to output neuron) & 0.02 \\
    maximum number of steps per episode & 1000 \\
\end{tabular}
\caption{Overview of the hyperparameters used for all experiments.}
\label{tab:hyperparameters}
\end {table}

\section{Using the Physical States for Training and Testing}
\label{sec:experimentsPhysicalStates}

\begin{figure}[ht]
\centering
\includegraphics[width=0.7\textwidth]{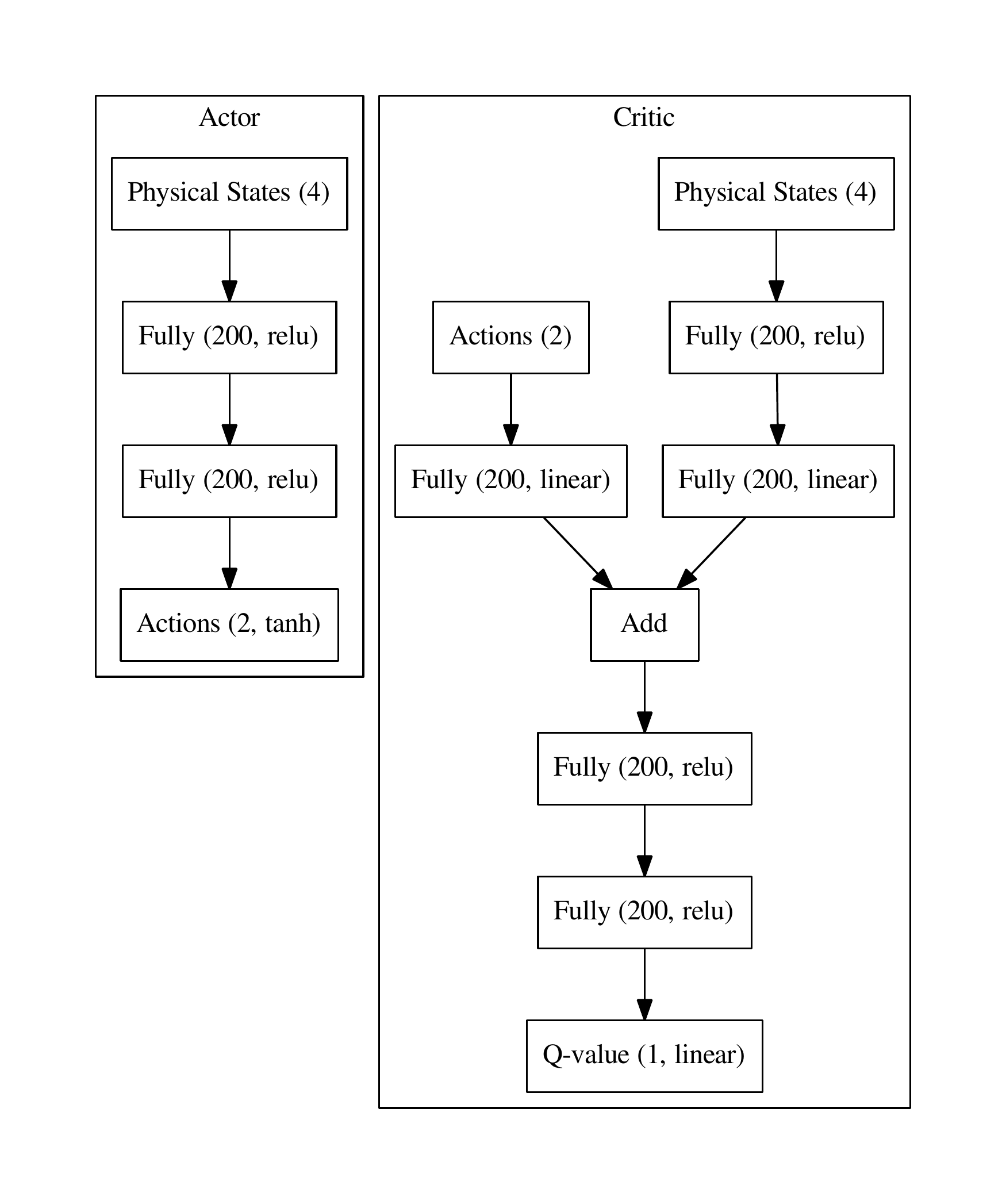}
\caption{Network structure used for all variants of DDPG to learn on the physical states. A weight penalty is added to the output neuron of the critic to prevent too fast rising Q-values.}
\label{fig:modelExperimentADdpg}
\end{figure}

At first, we compared our extended DDPG variants and the DDPG baseline, while learning directly from the physical states of the matplotlib simulation. This obviously requires the physical states also during test time. Figure \ref{fig:modelExperimentADdpg} depicts the network structure we used for the three experiments. We trained each algorithm for 1.600.000 steps and repeated the experiment 15 times. The asynchronous and distributed version both used 16 parallel threads. After each training phase, the trained model was tested for 100 episodes and the percentage of the successfully solved episodes was monitored. An episode is considered successful, when it takes less than 1000 steps for the gripper to reach the target. We state the mean and standard derivation for all independent tests and also the highest and the lowest of the 15 scores.

\begin{figure}[ht]
\centering
\includegraphics[width=0.47\textwidth]{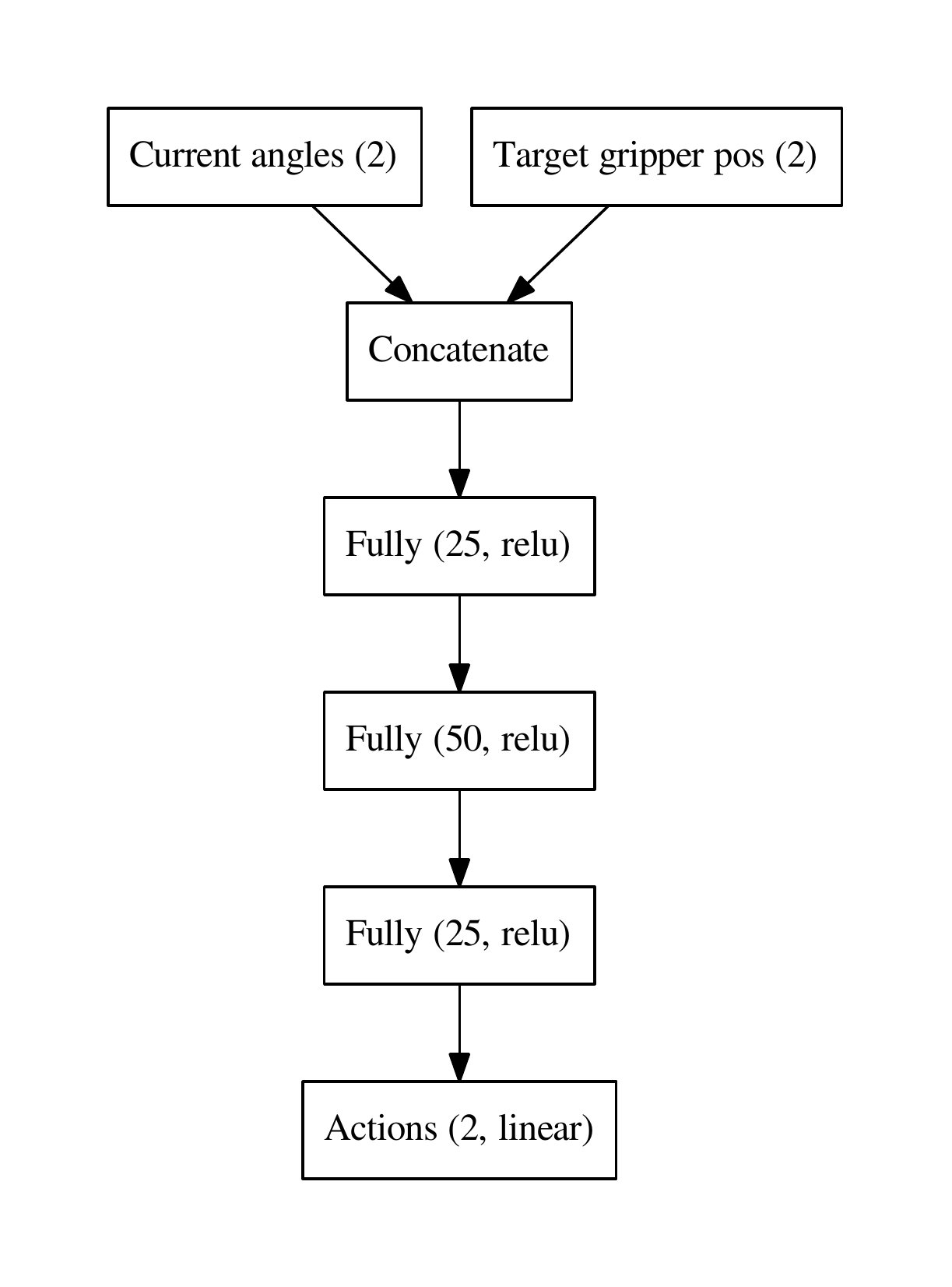}
\caption{Inverse model, which is trained to predict the action that moves the gripper to a target position, when the current angles of the arm segments are given.}
\label{fig:modelExperimentAInverse}
\end{figure}

In addition to the experiments using DDPG and its adapted variants, we also tested a simple inverse model, which predicts the action that is needed to guide the gripper of the robotic arm to a desired target position. This model was trained by simply executing random actions and training the model to output the executed actions while it observes the angles of the robotic arm at the state before the action was executed and the position of the gripper after the execution. The inverse model, that is shown in figure \ref{fig:modelExperimentAInverse}, was trained on 250.000 batches of size 1.000 drawn from an experience replay memory consisting of 1.000 episodes with random starting conditions and length 1.000 each. During evaluation we trained the inverse model three times from scratch and then tested each of these models 5 times on 100 episodes to obtain an evaluation similar to the DDPG experiments. The inverse model then does not receive an imagined next position of the gripper, but the true target position, that is most of the times out of reach.

\begin{table}[ht]
\centering
\begin{tabular}{ c|c|c|c|c }
 Experiment & Score Mean & SD & Max & Min \\
 \hline
 inverse model & 98.20\% & 1.28\% & 100.00\% & 97.20\% \\
 DDPG baseline & 49.40\% & 45.49\% & 100.00\% & 0.00\% \\
 distributed DDPG & 68.60\% & 12.53\% & 86.00\% & 46.00\% \\
 asynchronous DDPG & 74.67\% & 22.30\% & 96.00\% & 9.00\%
\end{tabular}
\caption{Summary of the experiments conducted using only the physical states. The inverse model performs best, while the asynchronous DDPG algorithm outperforms the baseline and the distributed version.}
\label{tab:experimentsPhysicalStates}
\end{table}

The results as depicted in table \ref{tab:experimentsPhysicalStates} show that asynchronous DDPG outperforms both other variants comparing the mean scores and is able to replicate the best results of the DDPG baseline. Although the asynchronous DDPG algorithm uses Hogwild! style updates, that only approximate the true gradient and induce a chance of one thread overriding the updates of another thread, this method seems to improve generalization and eventually yields better test results. Both modifications of DDPG reduce the training variance and do not depend on the starting conditions as heavily as the baseline. The distributed DDPG algorithm has very low variance, but is not able to replicate the best results of the baseline.

\begin{figure}[ht]
\centering
\includegraphics[width=0.7\textwidth]{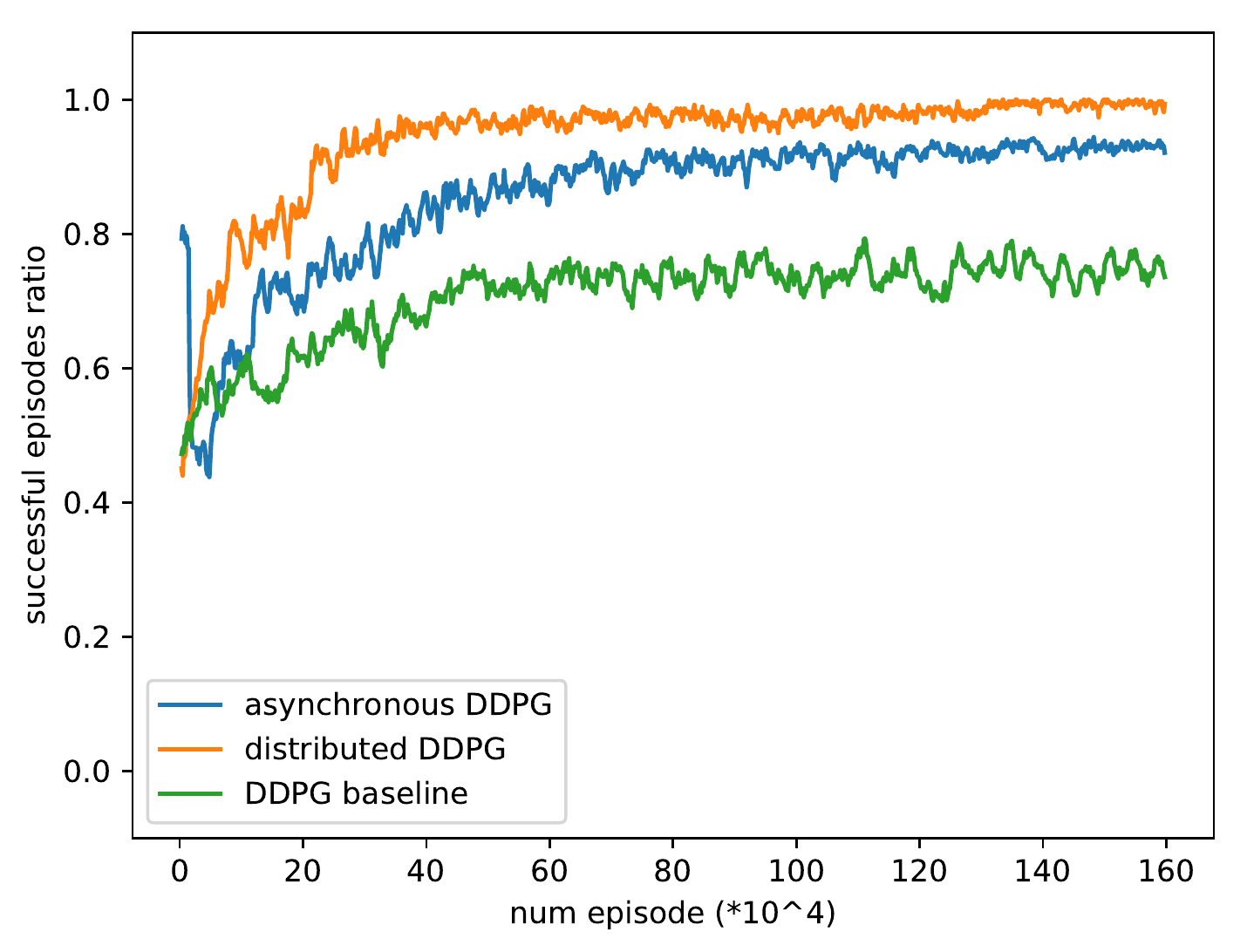}
\caption{Comparing asynchronous DDPG and distributed DDPG to the DDPG baseline. The plots state the relative amount of successful episodes over the last 25 episodes at the respective time. Every algorithm was trained 15 times for 1.600.000 steps and the scores were averaged.}
\label{fig:episodesPhysicalstates}
\end{figure}

Figure \ref{fig:episodesPhysicalstates} compares all variants of DDPG and shows the relative amount of successfully completed episodes during training. The modified DDPG algorithms both outperform the baseline and usually converge fast to better scores. The distributed DDPG algorithm converges best but lacks generalization as the test scores above showed. We also demonstrated that asynchronous training and DDPG can be combined in general to obtain a working algorithm. The training of our asynchronous DDPG implementation is more than five times faster than the DDPG baseline, because of the parallelization.

The best results however were obtained by using the inverse model, which is very stable and achieves 100\% success rate most of the time. The inverse model was solely trained on small movements of the robotic arm. Each arm segment is allowed to maximally move by 2 degrees in any direction, while the true target that is used as desired position for the gripper is often far away and requires many succeeding actions to be reached. The inverse model is still able to predict very good actions at every time step, although most of the data seen during testing has certainly never been observed during training. We like to see this result as evidence, that the inverse model is able to generalize very well and learns a good understanding of the system dynamics. Because of its very good performance we also tested the inverse model on the physical states of the Dart environment (see section \ref{sec:dartEnvironment}), where one inverse model was trained and tested three times for 500 episodes. The results are comparable to those before. We achieved an average score of 99.00\% with a standard derivation of 0.28\%. These findings demonstrate the ability of the inverse model to learn dynamics under different physical conditions.

\section{Using the Physical States for Training and Pixel Images for Testing}
\label{sec:experimentsPhysicalStatesForTraining}

We tested different options to include the physical states in the training process. As pixel images should be used later, it is required to base the reinforcement learning process on a representation that can be directly derived from pixels. We use the physical states during training to regularize a pretraining process, that learns to extract useful information from the pixels. In a second step, we use reinforcement learning to predict actions based on the outputs of the pretrained model. In the following, we will describe three different ways to extract information from pixels:

\begin{figure}[ht]
\centering
\includegraphics[width=0.4\textwidth]{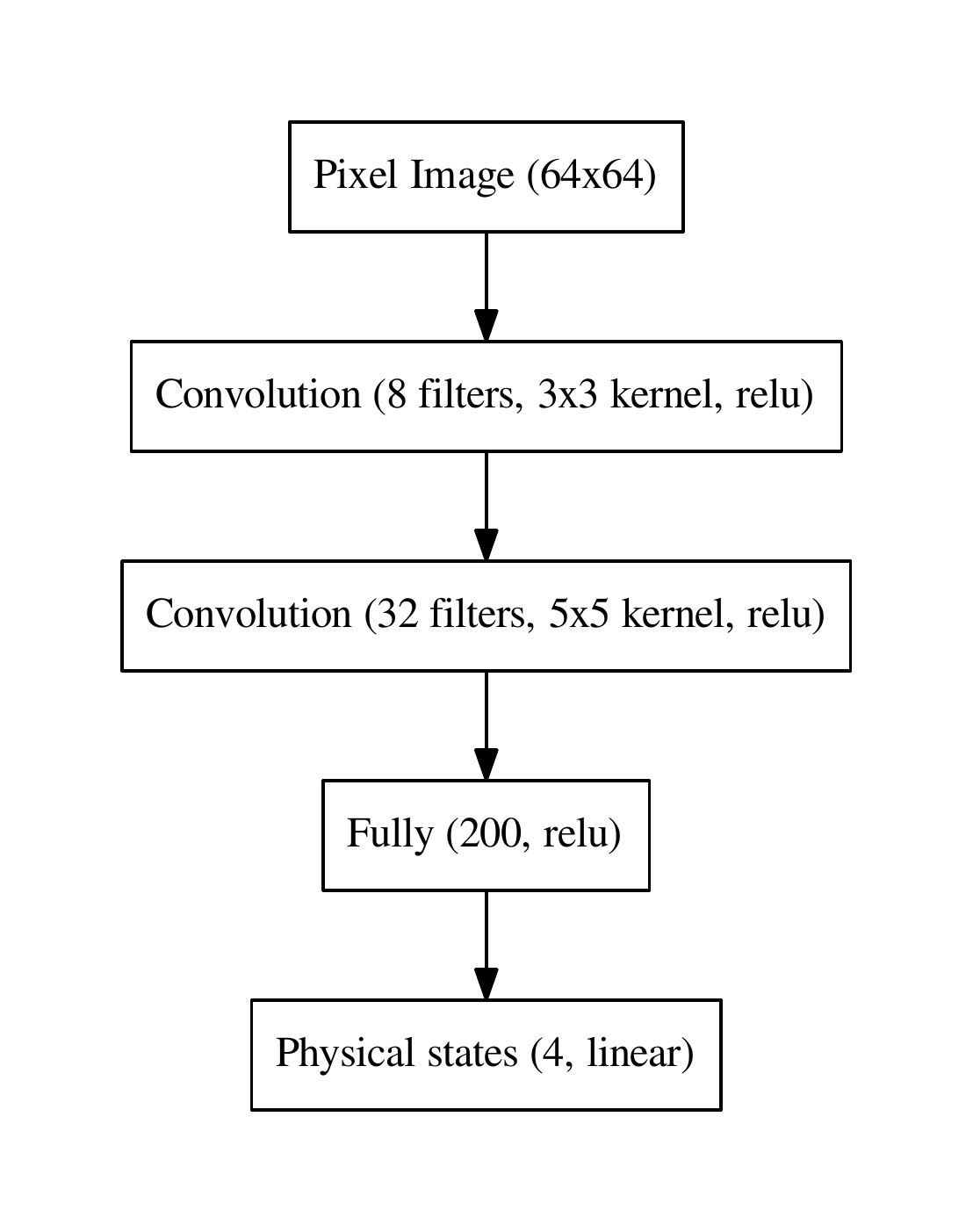}
\caption{Detailed architecture of the internal model, that consists of two convolutional layers and one fully connected hidden layer. }
\label{fig:internalModel}
\end{figure}

\begin{figure}[ht]
\centering
\includegraphics[width=0.68\textwidth]{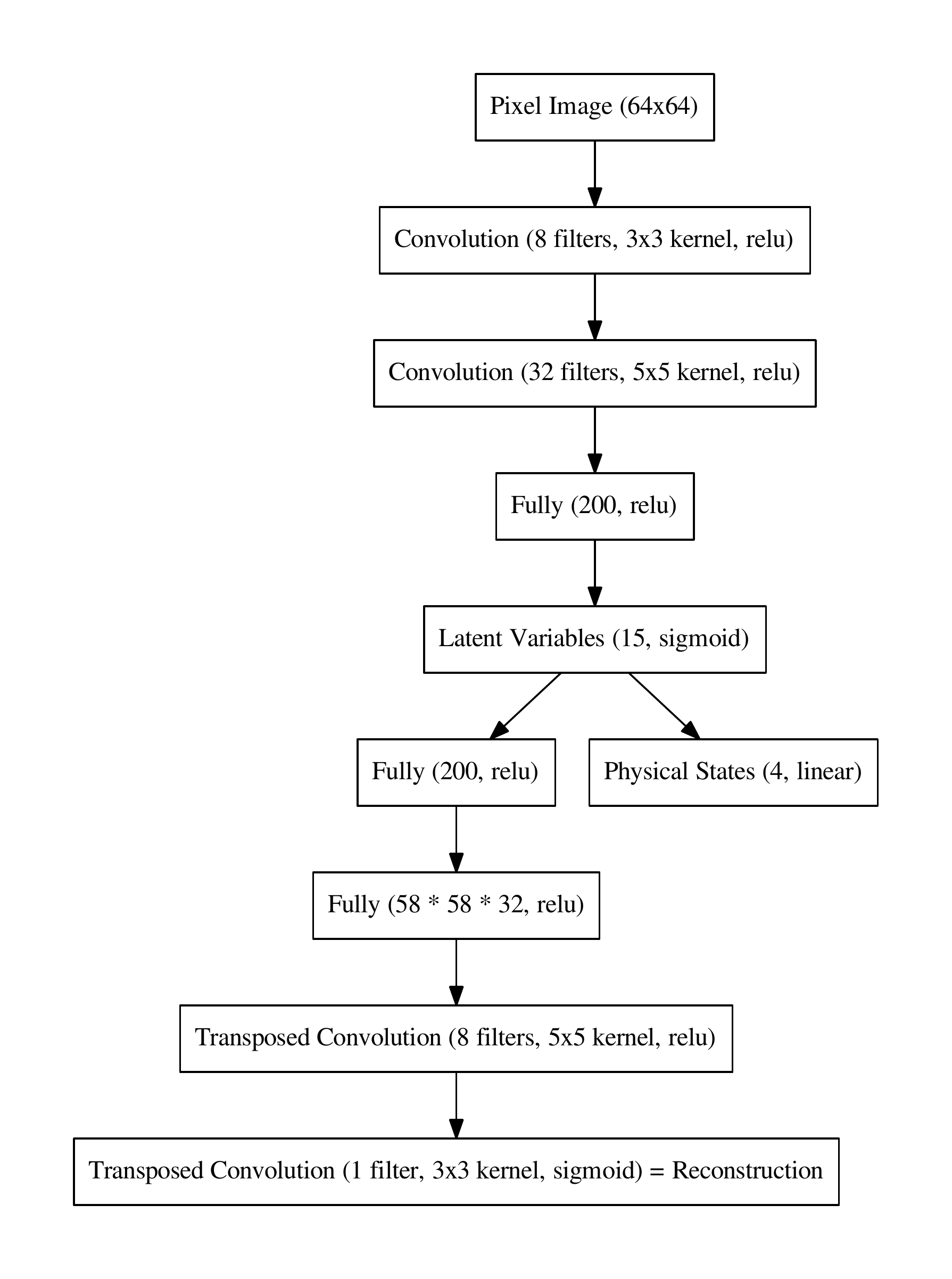}
\caption{Detailed architecture of the extended autoencoder, that also includes the physical states. We suppose the latent code of the autoencoder to become more useful, when the training process forces the physical states to be a linear function of it. The latent code is used as state input to the reinforcement learning algorithm. }
\label{fig:modelExperimentE}
\end{figure}

\begin{figure}[ht]
\centering
\includegraphics[width=0.5\textwidth]{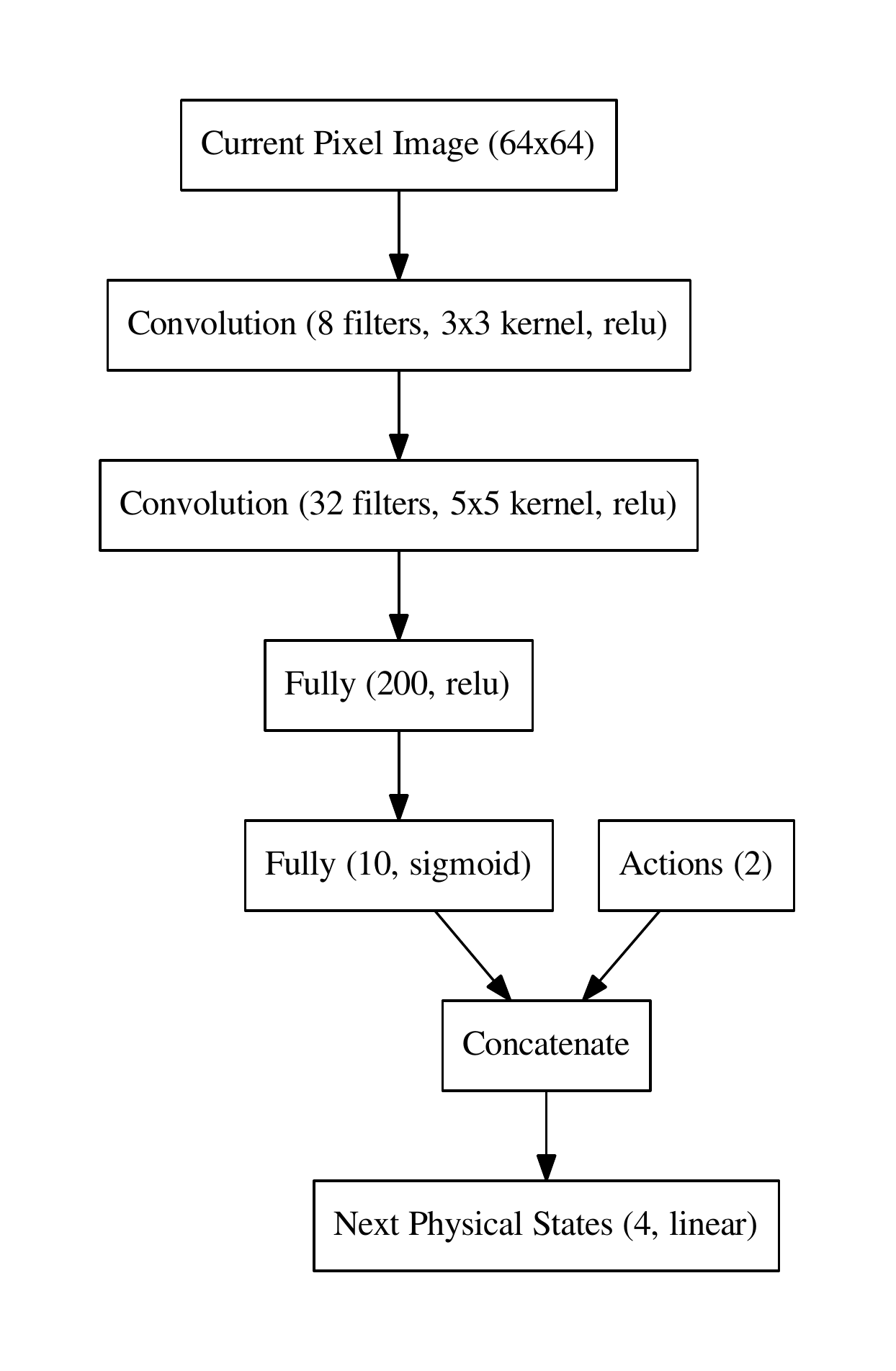}
\caption{Detailed architecture of the simple forward model. Learning the dynamics of the underlying physical system is expected to help finding useful features. The last layer, that only depends on the current pixels (sigmoid layer with 10 units) is fed to the reinforcement learning algorithm and can be predicted without knowledge of the physical states or actions. }
\label{fig:modelExperimentF}
\end{figure}

\textbf{1. Internal model:} The easiest way to make use of the physical states is pretraining a model that directly predicts the physical states from pixels, which we call an internal model. The concept has been described in section \ref{sec:pretrainingPhysicalStates}. The detailed internal model is shown in figure \ref{fig:internalModel}.

\textbf{2. Autoencoder, that additionally predicts physical states:} An autoencoder is normally used to simply reconstruct images. We extended an autoencoder structure to also predict the physical states like described in section \ref{sec:pretrainingAutoencoder}. The detailed network structure of the extended autoencoder is depicted in figure \ref{fig:modelExperimentE}. The physical states as additional training target are supposed to help the autoencoder finding more useful features and better encoding the pixel data.

\textbf{3. Forward model with physical states as output:} We trained a simple forward model (see section \ref{sec:inverseForwardModels} for comparision), that predicts the physical states of the following system state given the pixel image of the current state and the executed action in between. The last layer, that only depends on the pixel image of the current state, is used as state representation for the reinforcement learning algorithm. By forcing the model to learn about the dynamics of the system, this learned representation is expected to extract useful features for reinforcement learning. Figure \ref{fig:modelExperimentF} states the structure of the forward model.

The evaluation results of the internal model, the autoencoder with physical states and the forward model with physical states are given in table \ref{tab:experimentsHybrid}. We tested every method with a static background included in all images and also without a background. Every model was trained two times on 100.000 batches of size 250, that were randomly sampled from a memory consisting of 1.000 episodes with 1.000 steps each. All actions executed during the episodes were randomly sampled. We used the DDPG baseline for evaluation and ran reinforcement learning three times for each pretrained model. This means, we conducted six evaluations for every combination of model architecture and background, as each model was pretrained twice. The actor-critic architecture used for reinforcement learning is the same like depicted in figure \ref{fig:modelExperimentADdpg}, while only the size of the state input has been adapted for the autoencoder and the forward model. Every final actor was tested for 100 episodes and the relative amount of successfully completed episodes was recorded. As the variance for all experiments is high and due to long training time only few independent experiments could be performed for one pretrained model, we only state the maximum scores for comparision in table \ref{tab:experimentsHybrid}. The raw scores are listed in appendix \ref{ch:rawScores}.

\begin{table}[ht]
\centering
\begin{tabular}{ c|c|c }
 Experiment & background enabled & Max. Score \\
 \hline
Internal model & yes & 73\% \\
Internal model & no & 63\% \\
Autoencoder w. physical states & yes & 62\% \\
Autoencoder w. physical states & no & \textbf{90\%} \\
Autoencoder w. physical states & yes (mean removal) & 89\% \\
Forward model w. physical states & yes & 60\% \\
Forward model w. physical states & no & 82\% \\
\end{tabular}
\caption{Summary of the experiments conducted using the physical states during training and only pixel data for testing. }
\label{tab:experimentsHybrid}
\end{table}

The best results are obtained by using the autoencoder that predicts both images and physical states, when no background is included. The latent variables of the autoencoder thus provide a representation of the system state that is less error-prone than the physical states. The same argumentation applies to the performance of the forward model, which also outperforms the internal model in the case, when no background is included. The autoencoder gets heavily distracted by additive noise in the background of the images and the maximum score decreases. We therefore trained another version of this kind of extended autoencoder using noisy images, but performing mean removal before further processing them. This approach effectively removed most of the noise and again increased the score, which supports the assumption, that generating images with background noise is comparatively hard.

The additive noise makes image generation harder for the autoencoder, but has a weaker impact, when images are only used as input and convolutional layers can be used to remove the noise. For the inverse model, the scores for images with background are even a little bit higher.

\section{Using Pixel Images for Training and Testing}

\begin{figure}[ht]
\centering
\includegraphics[width=0.63\textwidth]{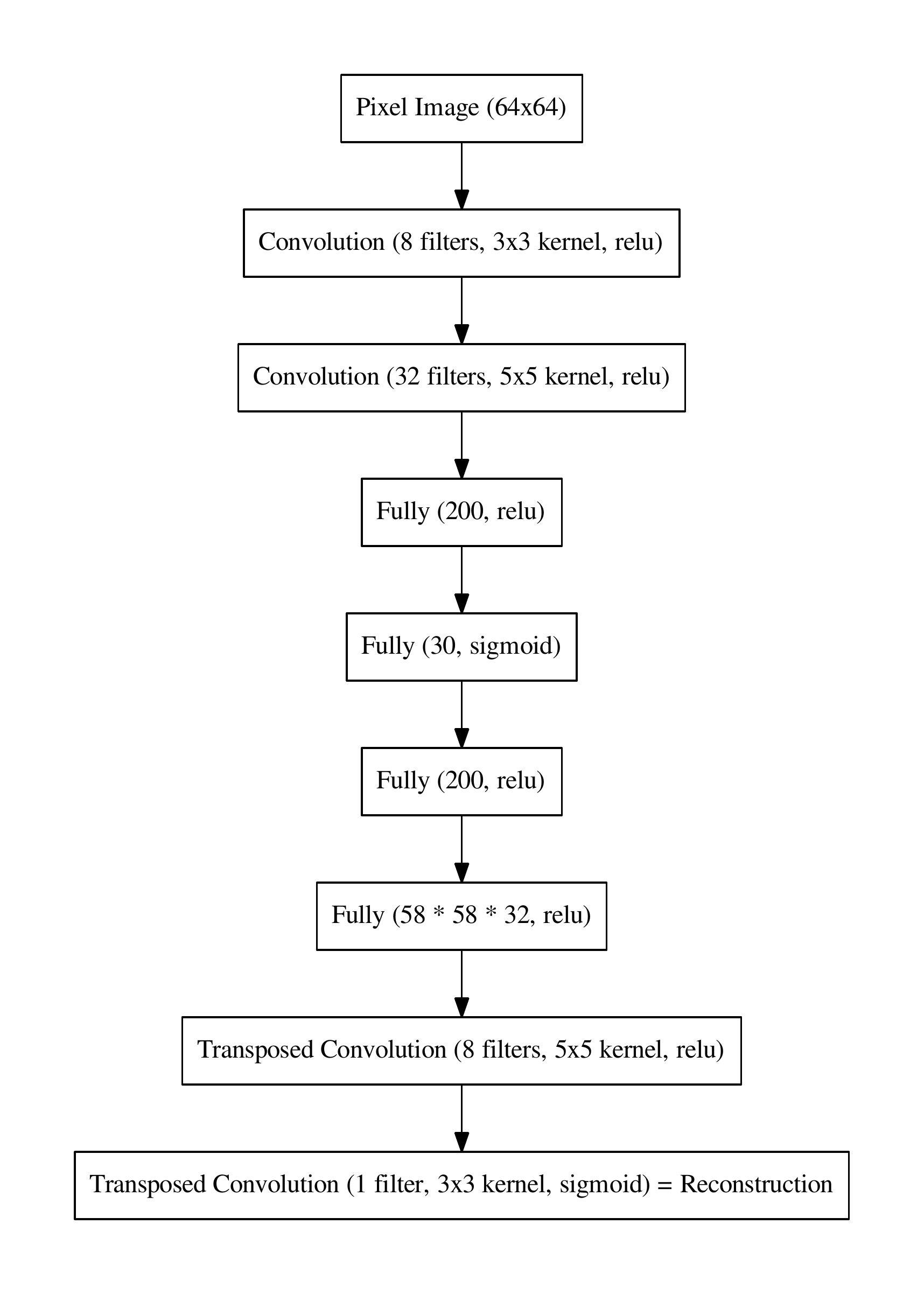}
\caption{Detailed architecture of the autoencoder, that simply reconstructs pixel images. While the model converges very well, it does not provide a useful state representation in the latent code and reinforcement learning fails to find a good policy. }
\label{fig:fullAutoencoder}
\end{figure}

\begin{figure}[ht]
\centering
\includegraphics[width=\textwidth]{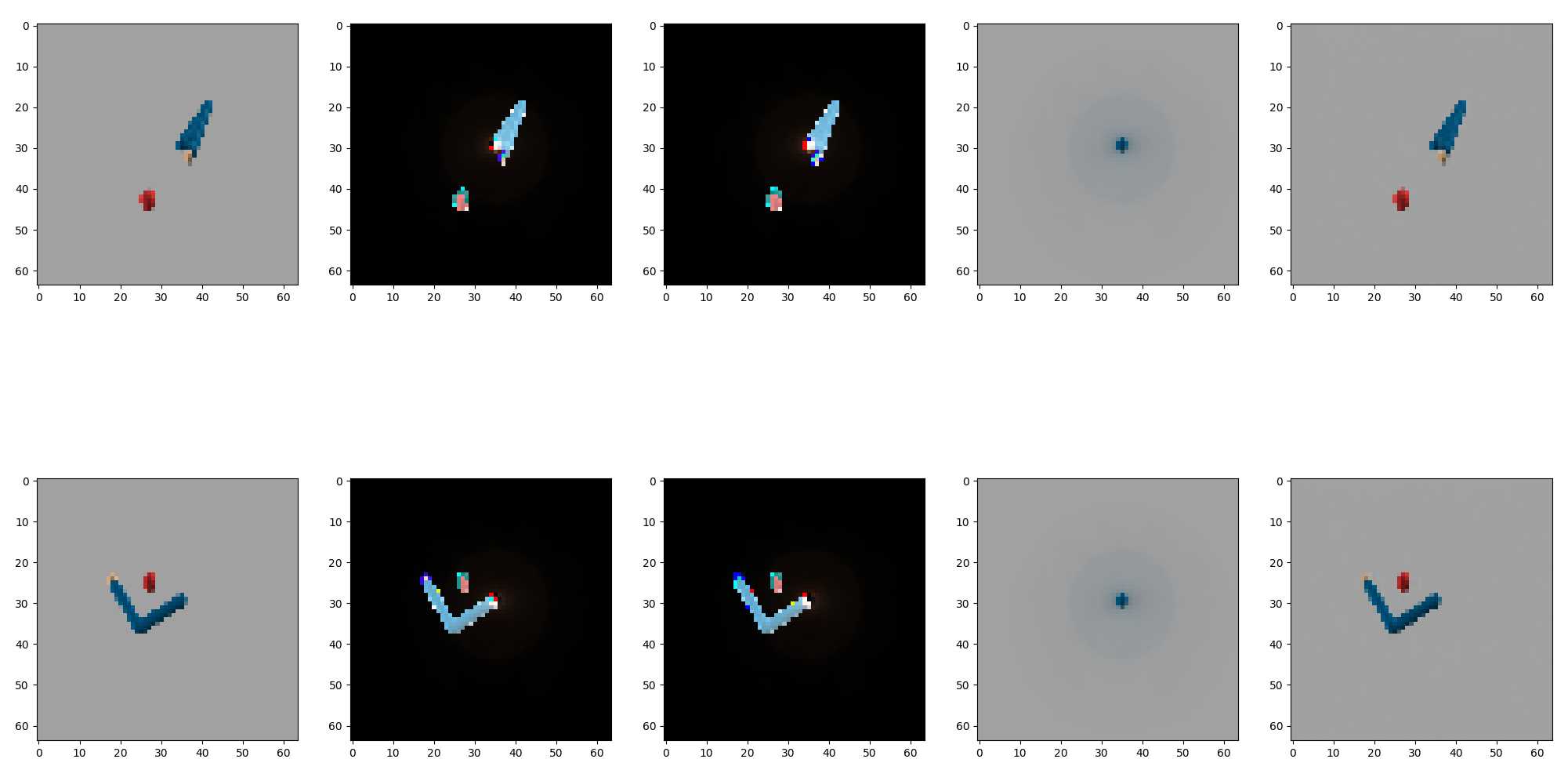}
\caption{Input to the converged autoencoder and reconstructed images, exemplary shown for the Dart environment: We apply mean removal to all images before passing them to the autoencoder and add the mean to the reconstructed images. The reconstruction looks very similar to the input. The two rows represent two different random inputs. The columns from left to right depict the following: 1 - input image, 2 - input image minus mean image, 3 - reconstruction of the autoencoder, 4 - mean image, 5 - reconstruction plus mean image. }
\label{fig:fullAutoencoderPrediction}
\end{figure}

We trained an autoencoder structure, that is shown in figure \ref{fig:fullAutoencoder} with the mean-squared-error function and mean removal to reconstruct pixel images. The trained autoencoder was able to reconstruct the input images almost perfectly for the matplotlib environment and even for the more sophisticated Dart simulation, as depicted in figure \ref{fig:fullAutoencoderPrediction}, with only minor modifications of the architecture to being able to process colored images. Reinforcement learning using the latent code of this autoencoder however produced poor results. The best of six independently trained actor-critic architectures, that were trained using the latent code of converged autoencoders as input and tested for 100 episodes each, was only able to solve 25\% of all episodes. We trained the autoencoder two times on the matplotlib environment and ran reinforcement learning three times for each pretrained model to obtain an evaluation similar to section \ref{sec:experimentsPhysicalStatesForTraining}. A fully random policy might also be able to achieve that score by randomly hitting the target sometimes. We also trained a variational autoencoder by replacing the 30 latent variables with 2x30 neurons that model the variance and mean of a multivariate gaussian distribution with 30 variables. The variational autoencoder did not converge. Furthermore, we tested several variants of inverse- and forward models only on pixels, but these did not help to improve the previously obtained score, or also completely failed to converge.

We were however able to use our implementation of the asynchronous DDPG algorithm to obtain a converging end-to-end learning algorithm. The architecture of the actor-critic network is very similar to figure \ref{fig:modelExperimentADdpg}. Two convolutional layers were added to process the visual information both for the actor and the critic. These layers replace the input of the physical states. Because of the long training time, when the gradient must be backpropagated through many additional weights, that are added with the convolutional layers, we only trained the end-to-end model once and tested it for 500 episodes. The model was able to solve 87\% of all episodes. The distributed DDPG algorithm and the DDPG baseline repeatedly failed the end-to-end learning task while using the same hyperparameters.

\chapter{Discussion}
\label{ch:discussion}

After describing the main concepts of deep reinforcement learning, we introduced two novel algorithms, that combine the deep deterministic policy gradient (DDPG) with asynchronous methods in different ways. We first evaluated these algorithms using a simple robotic task, while the true physical states of the environment were given, and compared both to a DDPG baseline. The most important finding of these experiments is, that the variant of DDPG using asynchronous lock-free gradient updates generalizes better than the variant with locks and also converges more often than a DDPG baseline, when executing multiple runs. The decision to use a lock free approach as introduced by \textcite{mnih2016asynchronous} was mostly motivated by performance considerations and not compared to a variant without locks. We suggest that lock-free updates might be beneficial not only to shorten training time but also for improving the generalization of many algorithms using asynchronous updates. We also showed, that the combination of DDPG and asynchronous updates can be applied to solve an end-to-end learning task. Another advantage of our asynchronous DDPG implementation is, that it is about five times faster than the DDPG baseline due to the parallelization.

We also investigated the effect of different pretraining techniques and successfully implemented multiple forms of pretraining, that use the physical states of the environment during training, but do not require them for testing. \textcite{levine2016end} show that real world robotic tasks sometimes provide access to the true physical states during training, but later require the trained model to act, while only observing camera images. The hybrid pretraining approach is much faster than reinforcement learning on pixel data, because pretraining is a standard deep learning task, that can be heavily parallelized and performed entirely on the GPU. In contrast to deep reinforcement learning, where it is necessary to regularly call the environment and thus communicate with the CPU, large data files can be prepared, which only need to be loaded once into memory. The simplified reinforcement learning task that follows after pretraining is also much faster than reinforcement learning on pixels, as the preprocessed state vector is comparatively small and the actor-critic model thus is far less complex.

The conducted experiments show, that it is more helpful to train a custom state representation than just predicting the physical states. This can be accomplished by adding an image reconstruction target similar to that of a deep autoencoder or modeling the system dynamics with inverse-/forward models. Referring to \textcite{agrawal2016learning}, we were able to show for a very simple experiment, that an inverse model can generalize well, while learning the system dynamics (see section \ref{sec:experimentsPhysicalStates}). An important question, that still remains open with our work, is how this generalization effect can be transfered to effectively process pixel data.

The internal model of section \ref{sec:experimentsPhysicalStatesForTraining} performs better on images with background noise, which at first might seem irrational. While this effect could be random variance and caused by the small number of collected scores, noisy images in fact sometimes proved useful for learning to detect features. \textcite{vincent2008extracting} showed that it can be beneficial to include additional noise in input images, because the trained model needs to find ways to distinguish noise from important features and thus learns a better representation of the important variations in the images. The same effect might apply here. For both other architectures, noisy images negatively influence the test scores. Learning to reconstruct images or learning the system dynamics presumably improves the detection of features in other ways and thus adding noise has an impact that is contrary to the positive effect we observed with the internal model. It might still be interesting to investigate the ability of all pretrained models to benefit from background noise, for example by using noisy images as input to the autoencoder and images without noise as reconstruction target. This is the principle of the denoising autoencoder (\cite{vincent2008extracting}).

 In section \ref{sec:experimentsPhysicalStatesForTraining} we also showed, that a major problem of pretraining pixel based models, is the need to generate pixel images. This is for instance the case when using an autoencoder network structure. \textcite{agrawal2016learning} showed that a combined inverse-/forward model can generalize well on pixel data, while they sidestep the complicated problem of generating images by first transforming the pixel images to a learned latent representation. Thereby, they establish a model, that combines the advantages of autoencoders with those of inverse- and forward-models. During our experiments, we tested both concepts independently. The convolutional layers, that carry out the transformation to the latent space are jointly trained with the combined inverse-/forward model. This joint training approach cannot be applied to our experiment, because the model only learns to detect objects in the images it is able to physically interact with. The arm of our simulated reacher task however does not physically interact with the target, which thus would have never been recognized.

Both of our simulated reacher experiments are similar to the OpenAI Gym reacher task\footnote{\url{https://gym.openai.com/envs/Reacher-v1/}, last downloaded 2017-09-14}, which we did not directly use, as it requires the proprietary mujoco library. The Dart environment has been compared to the OpenAI Gym reacher task and it has been shown, that learned policies can be transfered between the two environments\footnote{\url{https://github.com/DartEnv/dart-env/wiki/OpenAI-Gym-Environments}, last downloaded 2017-09-14}. We however concentrated on the simple matplotlib simulation for the most experiments, which makes learning arguably easier, because it does not include realistic physics and the visualization is very simple even when additional noise is added. The experiments we conducted are therefore not very comprehensive. For comparing to state-of-the-art methods in the field of deep reinforcement learning, most researchers evaluate their algortihms on many video games and simulated robotic tasks. It has become a quasi-standard to state the scores for all Atari games and use the mujoco simulations for robotic experiments (\cite{mnih2016asynchronous}; \cite{lillicrap2015continuous}). An interesting next step would therefore be to test the two novel variants of DDPG on a range of these experiments to being able to compare the respective scores and reliably assess the quality of these algorithms, while we showed that both algorithms can learn reasonable policies. Our distributed and asyncronous variants of DDPG are mainly motivated by the A3C algorithm. Both A3C and DDPG enjoy a high popularity and the underlying concepts are still used and further refined (\cite{odonoghue2016combining}; \cite{gu2016q}) or applied to various real world tasks (e.g. \cite{gu2017deep}). Hence, we think that deterministic policies and asynchronous learning in general still provide a good starting point for future research.

The DDPG baseline we used has very high variance, but was still used for all experiments with pretrained models, because we did not want to mix up the effects of the modifications to DDPG with the effects of pretraining a state model. We observed that the variance increased when we did not train on the physical states directly, but on an intermediate representation obtained by executing a pretrained model. This can be caused by the fact, that we used two different pretrained models of the same kind for all experiments in section \ref{sec:experimentsPhysicalStatesForTraining} and one model probably converged better than the other. Because the whole training process with pretraining and repeatedly predicting the intermediate representation during reinforcement learning is already relatively slow, we also collected less data than for the experiments on physical states (see section \ref{sec:experimentsPhysicalStates}). We also suppose, that learning on the intermediate representation is in general harder and thus the DDPG baseline fails more often. For the end-to-end learning algorithm, we were only able to train one model with asynchronous DDPG as training the convolutional layers was still very slow, despite the use of asynchronous updates.

We were generally able to provide a good solution with a score above 85\% for all three main categories of experiments: reinforcement learning directly on the physical states, pretraining a state model using the physical states supplementary to pixels and learning only from pixels. An unsuccesful approach was to pretrain a state model only from pixel data. \textcite{goodfellow2016deep} state, that unsupervised pretraining might be outdated for many applications, where end-to-end learning is possible. The reinforcement signal is supposed to provide helpful information to the algorithm and thus enhances the detection of features in images, that are useful for the task to solve. Purely unsupervised pretraining like training an autoencoder does not have this information and is therefore less efficient. Recent innovations like the variational autoencoder detect features in images very well, but are designed to generalize to small changes in position, orientation or shape of objects (\cite{doersch2016tutorial}). Robotic applications however often require exact knowledge of positions of objects or angles like in our simulated example (see section \ref{sec:trainingEnvironments}). Nevertheless, it is possible to still make use of inverse- or forward models, as they require the model to learn the dynamics of the system. \textcite{dosovitskiy2016learning} use a forward model to solve a control task without reinforcement learning, but only focus on discrete actions. We suppose that under these conditions, unsupervised pretraining can work, but might be more useful in other scenarios with different environements than ours. The model capacity of our tested models for unsupervised pretraining should be sufficient, as a very similar model was able to solve an end-to-end learning task, but regularization strategies like batch normalization (\cite{ioffe2015batch}) or others might be added. Mean removal seems to be crucial for the success of all models, that need to generate images. We also think, that the success of the simple inverse model of section \ref{sec:experimentsPhysicalStates} is mainly caused by its very fast training speed and the ability to train it on many millions of example transitions in few hours. It thus might be worthwhile to further investigate especially inverse- and forward models, that only have access to pixel data, and train them for a very long time.

None of our eperiments included any form of recurrence or memory in the network structure. Recurrent neural networks like the LSTM (\cite{hochreiter1997long}) have been successfully used by many researchers to solve reinforcement learning problems, where only parts of the environment can be observed and the whole process forms a POMDP (e.g. \cite{mnih2016asynchronous}). Partially observed environments were shortly mentioned in section \ref{sec:rlissues}, while we did not consider them in any experiment throughout this thesis. Nevertheless, we suppose that many of the used network structures can be augmented with recurrent layers and thus can also be applied to partially observable environments.

\printbibliography

\appendix
\chapter{Raw Scores}
\label{ch:rawScores}

\begin{table}[ht]
\centering
\begin{tabular}{ c|c|c }
 Experiment & background enabled & Raw Scores \\
 \hline
Internal model & yes & 73\%, 65\%, 62\%, 52\%, 48\%, 2\%  \\
Internal model & no & 63\%, 61\%, 55\%, 2\%, 1\%, 1\% \\
Autoencoder w. physical states & yes & 62\%, 59\%, 34\%, 28\%, 1\%, 0\% \\
Autoencoder w. physical states & no & 90\%, 3\%, 2\%, 2\%, 0\%, 0\% \\
Autoencoder w. physical states & yes (mean removal) & 89\%, 85\%, 82\%, 56\%, 1\%, 0\% \\
Forward model w. physical states & yes & 60\%, 60\%, 52\%, 0\%, 0\%, 0\% \\
Forward model w. physical states & no & 82\%, 50\%, 23\%, 1\%, 0\%, 0\% \\
\end{tabular}
\caption{Raw scores for the experiments conducted using the physical states during training and only pixel data for testing. }
\label{tab:rawScoresHybrid}
\end{table}

\begin{table}[ht]
\centering
\begin{tabular}{ c|c }
 Experiment & Raw Scores \\
 \hline
Full Autoencoder & 25\%, 12\%, 8\%, 5\%, 3\%, 2\%
\end{tabular}
\caption{Raw scores for the experiments conducted using pixel images during training and testing. }
\label{tab:rawScoresPixels}
\end{table}

\end{document}